%% file: main.tex
\documentclass{article}
\usepackage{arxiv,times}
\usepackage{placeins}

\input{math_commands.tex}

\usepackage{wrapfig}

\usepackage{subcaption}
\usepackage{amsmath,amssymb,bm,amsthm}
\usepackage{pifont}
\usepackage{booktabs}
\usepackage{xcolor}
\usepackage{hyperref}
\usepackage{url}

\usepackage{multirow}

\usepackage{graphicx}

\usepackage{arydshln}
\theoremstyle{plain}

\theoremstyle{definition}

\theoremstyle{remark}

\title{GS-MoE: Training-Free MoE Compression via \\Groupwise Subspace Sharing and Pruning}

\author{%
  Carson Cary\textsuperscript{1}, Yukai Jin\textsuperscript{1}, Yihao Huang\textsuperscript{1}, Zhankai Ye\textsuperscript{1}, \\
  \textbf{Fangyi Li\textsuperscript{2}, Ting Hua\textsuperscript{3}, Xin Liu\textsuperscript{1}, Shangqian Gao\textsuperscript{1}} \\[2pt]
  \textsuperscript{1}Florida State University \quad
  \textsuperscript{2}University of Pennsylvania \quad
  \textsuperscript{3}University of Notre Dame
}

\newcommand{\dff}{d_{\mathrm{ff}}}

\makeatletter
\newdimen\dcmrdash \dcmrdash=2pt   %
\newdimen\dcmrgap  \dcmrgap=2pt    %
\def\dcmr@leaders{{\CT@arc@\xleaders\hbox{\vrule\@height\@thisrulewidth\@width\dcmrdash\kern\dcmrgap}\hfill\kern\z@}}
\def\dcmr@restore{\global\let\@cmidrulea\dcmr@savea\global\let\@cmidruleb\dcmr@saveb}
\def\dcmr@a{\multispan\@cmidla&\multispan\@cmidlb\dcmr@restore
   \unskip\hskip\cmrkern@l\dcmr@leaders\hskip\cmrkern@r\cr}
\def\dcmr@b{\multispan\@cmidlb\dcmr@restore
   \unskip\hskip\cmrkern@l\dcmr@leaders\hskip\cmrkern@r\cr}
\newcommand\dashedcmidrule{\noalign{\global\let\dcmr@savea\@cmidrulea\global\let\dcmr@saveb\@cmidruleb
   \global\let\@cmidrulea\dcmr@a\global\let\@cmidruleb\dcmr@b}\cmidrule}
\makeatother

\DeclareMathOperator{\tr}{tr}

\iclrfinalcopy 
\begin{document}
\maketitle

\begin{abstract}
   \input{Sections/Abstract}

\end{abstract}

\input{Sections/Introduction}

\input{Sections/RelatedWork}

\input{Sections/Method}

\input{Sections/Results}

\input{Sections/Conclusion}

\bibliography{References}
\bibliographystyle{arxiv}

\appendix
\input{Sections/Appendix}

\end{document}

%% file: math_commands.tex
\usepackage{amsmath,amsfonts,bm}

\def\eqref#1{equation~\ref{#1}}

\def\1{\bm{1}}

\DeclareMathAlphabet{\mathsfit}{\encodingdefault}{\sfdefault}{m}{sl}
\SetMathAlphabet{\mathsfit}{bold}{\encodingdefault}{\sfdefault}{bx}{n}

\newcommand{\R}{\mathbb{R}}

%% file: Sections/Abstract.tex
Mixture-of-Experts (MoE) models reduce computation through sparse activation, but storing their expert parameters remains a major memory bottleneck.
We introduce GS-MoE, a training-free compression framework that addresses redundancy across and within experts through two compression axes: shared-subspace rank and expert intermediate width.
Groupwise low-rank representations capture shared structure across experts while retaining expert-specific projections, and Nystr\"om-based neuron compression reduces intermediate width within each expert.
For fixed shared subspaces, we derive per-expert reconstruction-error reductions from the reconstruction objective and use them to guide expert grouping.
Budgets are allocated hierarchically using a gradient-free layer-importance score to guide the allocation of rank and width budgets across layers, which eigenvalues and leverage then distribute as non-uniform ranks across groups and intermediate widths across experts.
Experiments demonstrate the complementary benefits of compression along both axes, particularly at high compression rates on fine-grained MoE models.
On DeepSeek-MoE-16B-Base, at a 60\% reduction in total model parameters, GS-MoE retains approximately 90\% of the uncompressed model's average reasoning performance, compared with 73\% for D$^2$-MoE, while achieving 20\% lower out-of-domain perplexity. We provide a code implementation at: https://github.com/cbacary/GS-MoE

%% file: Sections/Introduction.tex
\section{Introduction}
\label{intro}

Mixture-of-Experts (MoE) models increase parameter capacity while limiting computation by activating only a subset of experts for each token \citep{bib:DeepSpeed,bib:DeepseekAI}.
Sparse activation, however, does not eliminate the storage cost of inactive experts.
The resulting expert parameter footprint remains a major obstacle to deployment, motivating compression methods that reduce memory requirements while preserving model quality \citep{bib:MoE-SVD,bib:D2-MoE}.

Expert compression must account for both structure shared across experts and redundancy within each expert.
Shared-factor methods such as MoE-SVD address the former with a common low-rank factor across experts where D$^2$-MoE stores a shared base with compressed expert-specific residuals \citep{bib:MoE-SVD,bib:D2-MoE}.
These parameterizations impose different constraints on how shared and expert-specific information is represented.
In particular, MoE-SVD requires the corresponding expert factors to use one common subspace; directions outside that subspace cannot be recovered by changing the expert-specific projections alone.
Intermediate-neuron pruning addresses a different limitation by reducing expert width \citep{bib:TENP,bib:LessIsMoE}, but pruning alone still stores separate weight representations for the retained neurons of different experts.
Thus, a fixed shared subspace restricts where expert-specific variation can be represented, whereas width reduction alone does not encode common structure through shared parameters.
This motivates a representation that allows subspaces to vary across expert groups and intermediate widths to vary across experts, with both capacities controlled under a common parameter budget.

We introduce \textbf{GS-MoE}, a training-free framework that compresses MoE parameters along shared-subspace rank and expert intermediate width.
Group-shared projections represent common structure across experts, while expert-specific projections express individual transformations within each group's subspaces.
This makes the scope of parameter sharing explicit rather than fixing a single basis for the entire layer.
Ridge-leverage neuron selection reduces intermediate width before the up projections are factorized.
We estimate the shared up projections using a group-level surrogate and fit the expert-specific projections using each expert's input statistics.
The down reconstruction objective admits a closed-form solution before width compression; Nystr\"om reconstruction \citep{bib:NystromMatrix} then maps the retained hidden activations to the down projection's shared coordinates.
Group rank and expert width therefore control two dimensions of the same compressed representation: the capacity of shared subspaces and the size of expert-specific intermediate activations.

GS-MoE determines which experts share a subspace and how capacity is distributed under a parameter budget using calibration statistics.
For fixed neuron selections and candidate shared projections, fitting the expert-specific projections expresses the reconstruction objective as a constant minus per-expert error reductions, which guide expert grouping.
A gradient-free layer-importance score based on normalized, gate-weighted expert output magnitudes \citep{bib:REAP} guides rank and width budgets across layers.
Within each layer, neuron leverage scores determine expert widths, and eigenvalue-based scores and parameter costs guide group ranks.
The rank scores reflect the up surrogate and the down reconstruction objective before width compression.
This allocation adapts the compressed representation at the layer, group, and expert levels without post-compression training.

Experiments demonstrate the complementary benefits of compression along both axes, particularly at high compression rates on fine-grained MoE models.
On DeepSeek-MoE-16B-Base, at 60\% compression, GS-MoE retains 90\% of the uncompressed model's average reasoning performance, compared with 73\% for D$^2$-MoE, while achieving 20\% lower out-of-domain perplexity.
Our contributions are:
\begin{itemize}
    \item We introduce GS-MoE, a training-free framework that compresses expert parameters along shared-subspace rank and intermediate width, addressing shared structure across experts and redundancy within experts.
    \item We derive per-expert reconstruction-error reductions for fixed neuron selections and shared projections, providing an error-guided criterion for expert grouping.
    \item We allocate capacity across layers, groups, and experts using a normalized expert-output score, eigenvalue-based rank scores, and neuron leverage scores under a parameter budget.
    \item We demonstrate the benefit of compression along both axes and improved reasoning retention and out-of-domain perplexity over D$^2$-MoE at 60\% compression on DeepSeek-MoE-16B-Base.
\end{itemize}

%% file: Sections/RelatedWork.tex
\section{Related Work}
\label{related_work}

\subsection{Pruning and Merging of Mixture-of-Experts}

Through sparse activation the MoE architecture effectively limits computational demand but introduces significant parameter overhead and redundancy making MoE compression techniques vital.
These techniques generally operate through one of expert merging, expert pruning, neuron pruning, or decomposition.
Expert merging methods reduce parameters by consolidating similar experts: HC-SMoE \citep{bib:HC-SMoE} uses hierarchical clustering to merge similar experts without retraining while DERN \citep{bib:DERN} redistributes neuron-level segments from pruned experts to compatible retained experts before merging them.
Not All Experts are Equal \citep{bib:NAEE} introduces expert pruning which reduces parameter overhead through whole expert removal; REAP \citep{bib:REAP} builds on this by applying router-weighted output norms to score expert prunability.
Further, REAP \citep{bib:REAP} identifies that merging-based methods incur significant error that pruning-based methods do not.
Rather than discarding entire modules, finer-grained frameworks such as TENP \citep{bib:TENP} and Less-Is-MoE \citep{bib:LessIsMoE} compress intermediate expert widths, out-performing whole expert removal.
However, pruning techniques rely solely on parameter deletion, leaving shared inter-expert structures unexploited.
In contrast, our method avoids the distortion of expert merging and the finality of pure pruning by coupling group-shared subspaces that capture inter-expert redundancy with intra-expert width compression.

\subsection{Low-Rank Decomposition and Parameter Sharing}
Decomposition-based frameworks replace expert weight matrices with compact low-rank factors.
Early decomposition-based approaches apply low-rank approximation in conjunction with expert pruning \citep{bib:MoE-I2}.
Alternative methods employ parameter sharing to exploit cross-expert redundancy.
MoE-SVD \citep{bib:MoE-SVD} shares singular vectors from the most frequently activated experts and D$^2$-MoE \citep{bib:D2-MoE} represents each expert as a shared base weight plus a low-rank expert-specific residual.
TD-MoE stacks expert weights into a tensor for joint Tucker decomposition with adaptive three-dimensional rank allocation and multi-linear whitening \citep{bib:TD-MoE}.
In each of these three methods, the shared component, whether a common factor, an additive base, or a set of mode factors, spans every expert in the layer.
An expert whose weights diverge must absorb the mismatch in its expert-specific component; under MoE-SVD, directions outside the shared subspace cannot be recovered by any choice of expert-specific factors.
GS-MoE shares subspaces within groups of experts, and it forms those groups to maximize the reconstruction-error reduction each expert obtains, so experts with dissimilar input or output subspaces are not forced onto a common basis.
Additionally, GS-MoE adapts MoDeGPT \citep{bib:MoDeGPT}, an intermediate-width compression method for dense LLMs, to its MoE compression framework.
GS-MoE minimizes reconstruction error over width-reduced up projections, effectively folding an additional compression axis directly into the subspace.

Other methods relax the single-shared subspace constraint.
LorExperts~\citep{bib:LorExperts} keeps one full-precision dominant expert per
cluster and represents the other members as low-rank corrections to it. %
MoBE \citep{bib:MoBE} builds each expert's up and gate matrices from its own
combination of basis matrices shared across the layer, fitted by gradient-based
weight reconstruction, leaving the down projections uncompressed.
Sub-MoE \citep{bib:Sub-MoE} utilizes k-means clustering to group experts clusters by
cosine similarity, align clusters to a shared basis by SVD, then merge the clusters
into a single expert.
Our method uses balanced medoid clustering with $k$-means++ initialization to group experts using reconstruction-error reductions.
MoLAE \citep{bib:MoLAE} is closest to GS-MoE's reparameterization through its group-shared latent projections with expert-specific transformations via stacked SVD. However, MoLAE forms arbitrary groups, and is forced to preserve the up projections entirely and restrict intermediate width to the latent rank forming compression factors of $\R^{m \times m}$, bounding its compression by the ratio of the two dimensions. %
GS-MoE retains and compresses every expert, including its down projections,
and treats group rank and expert width as separate budgeted quantities
allocated across layers, groups, and experts from calibration statistics, so
its compression is set by the parameter budget rather than by a fixed
dimension ratio or an uncompressed component. %

%% file: Sections/Method.tex
\section{Method}
\label{sec:method}

\subsection{Notation}
\label{sec:notation}

We consider one layer with $E$ experts, model dimension $d$, and intermediate width $\dff$; the layer index $l$ is included only for layer-wise capacity allocation. In group-wise sharing, $e \in \mathcal E_g$ denotes that expert $e$ belongs to group $g$, with $G$ groups in total.
We denote the original expert weights by $W^{\star}_e$,
group-shared projection matrices by $P^{\star}_g$,
and expert-specific projection matrices by $A^{\star}_e$,
where $\star \in \{\uparrow,\downarrow\}$ indicates the up or down projections. For simplicity, we omit gated projections.

\subsection{Compression along Rank and Width}
\label{sec:representation}

An FFN expert computes $f_e(x)=W_e^\downarrow\phi(W_e^\uparrow x)$, where $\phi(\cdot)$ is the nonlinear operation, $W_e^\uparrow\in\mathbb R^{\dff\times d}$ and $W_e^\downarrow\in\mathbb R^{d\times\dff}$.
After compression with GS-MoE, we have the following computation for $e\in\mathcal E_g$:
\begin{equation}
    \label{eq:compressed_expert}
    \hat f_e(x)=P_g^\downarrow A_e^\downarrow
    \phi\!\left(A_e^\uparrow(P_g^\uparrow)^\top x\right).
\end{equation}
Here $P_g^\uparrow,P_g^\downarrow\in\mathbb R^{d\times r_g}$ are group-shared projections, where $r_g$ is the latent dimension of group $g$. The expert-specific projections $A_e^\uparrow\in\mathbb R^{k_e\times r_g}$ and $A_e^\downarrow\in\mathbb R^{r_g\times k_e}$ retain $k_e$ intermediate neurons for expert $e$. The input is first projected into the $r_g$-dimensional shared space by $(P_g^\uparrow)^\top$, then mapped to $k_e$ intermediate activations by $A_e^\uparrow$. After the nonlinearity, $A_e^\downarrow$ maps these activations back to $r_g$ dimensions, and $P_g^\downarrow$ restores the model dimension $d$. For gated experts, the gate and up projections use separate shared and expert-specific factors with the same $r_g$ and $k_e$. The gate projection follows the same derivation as the up projection.

\subsection{Groupwise Subspace Sharing and Pruning}
\label{sec:factorization}
\label{sec:nystrom_method}

\begin{figure}[h]
    \centering
    \includegraphics[width=0.9\linewidth]{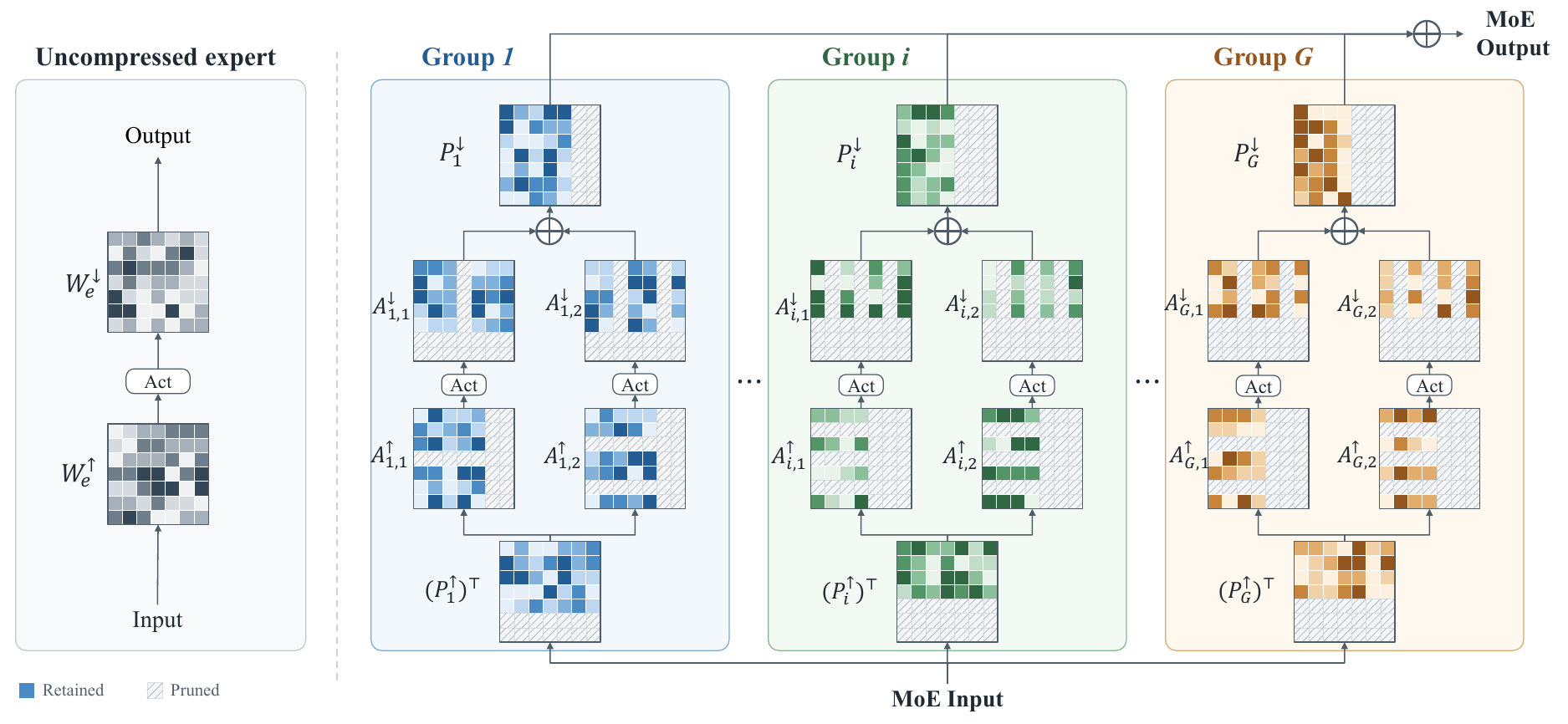}
    \caption{Overview of GS-MoE. Compared with an uncompressed expert (left), GS-MoE partitions experts into equally sized groups with shared input and output projections \(P_g^\uparrow\) and \(P_g^\downarrow\), while retaining expert-specific projections \(A_{g,j}^\uparrow\) and \(A_{g,j}^\downarrow\), where \(j\) indexes experts within group \(g\), so \(A_{g,j}^{\star}\) corresponds to \(A_e^{\star}\) for the associated expert \(e\in\mathcal E_g\). Group ranks and expert intermediate widths are allocated non-uniformly under a fixed parameter budget. Colored cells indicate retained parameters, and hatched regions indicate removed dimensions. Two experts per group are shown for illustration; routing weights and gated projections are omitted.}
    \label{fig:method_figure}
    \vspace{-10pt}
\end{figure}

We estimate shared projections and expert-specific projections separately, with neuron selection applied before up factorization and reconstruction applied after down factorization.
The up projection requires a group-level surrogate, whereas the down projection admits an exact solution for its corresponding reconstruction objective before width compression.

\textbf{Calibration and neuron selection.}
For each expert $e$, let $h_e$ be its original intermediate hidden activation and $y_e=W_e^\downarrow h_e$ its output.
We collect the second moments $C^x_e=\mathbb E_e[xx^\top]$, $C^h_e=\mathbb E_e[h_eh_e^\top]$, and $C^y_e=\mathbb E_e[y_ey_e^\top]$, where $\mathbb E_e$ averages over calibration inputs routed to expert $e$.
Given total width $K=\sum_e k_e$, we select neurons by their ridge-leverage scores
$q_{e,i}=[C^h_e(C^h_e+\eta I)^{-1}]_{ii}$, where $\eta>0$.
The top-$K$ scores across experts determine $k_e$ and the selection matrix $S_e\in\mathbb R^{\dff\times k_e}$, whose columns are the corresponding columns of the identity matrix.
The same $S_e$ selects the up and gate neurons.

\textbf{Up projection:}
For fixed neuron selections, the up projection minimizes the reconstruction error of the selected intermediate pre-activations:
\begin{equation}
    \label{obj:up_objective}
    \min_{P_g^\uparrow,A_e^\uparrow}
    \sum_{e\in\mathcal E_g}\mathbb E_e
    \left\|S_e^\top W_e^\uparrow x-A_e^\uparrow(P_g^\uparrow)^\top x\right\|_2^2,
    \quad (P_g^\uparrow)^\top P_g^\uparrow=I_{r_g}.
\end{equation}
The joint up problem does not, in general, admit a closed-form solution when expert input statistics differ.
Although the expert-specific projections have closed-form solutions for a fixed shared projection, eliminating them leaves expert-dependent inverse terms that still depend on the shared projection.
We therefore estimate the shared projection using a group-level surrogate and then fit $A_e^\uparrow$ using each expert's own input statistics.
Let $C_g=\sum_{e\in\mathcal E_g}\tau_e C^x_e$, where $\tau_e=n_e/\sum_{j\in\mathcal E_g}n_j$ and $n_e$ counts routed calibration assignments.
Using the surrogate matrix $H_g^\uparrow$ defined in Section~\ref{sec:allocation}, the resulting shared and expert-specific projections are
\begin{equation}
    \label{eq:up_solution}
    \begin{aligned}
        P_g^\uparrow & =\operatorname{orth}\!\left(C_g^{-1/2}
        \operatorname{eig}_{r_g}(H_g^\uparrow)\right),          \\
        A_e^\uparrow & =S_e^\top W_e^\uparrow C^x_eP_g^\uparrow
        \left((P_g^\uparrow)^\top C^x_eP_g^\uparrow\right)^{-1}.
    \end{aligned}
\end{equation}
Here $\operatorname{eig}_r(\cdot)$ returns the leading $r$ orthonormal eigenvectors, and $\operatorname{orth}(\cdot)$ orthonormalizes the columns.
The expert-specific projections minimize the error for the chosen shared projection; the shared projection is an approximation to the original objective.
The surrogate construction and expert-specific projection solution are derived in Appendix~\ref{app:up_solution}.
The gate projection follows the same construction.

\textbf{Down projection:}
The down projection minimizes the reconstruction error of the original expert outputs:
\begin{equation}
    \label{obj:down_objective}
    \min_{P_g^\downarrow,B_e}
    \sum_{e\in\mathcal E_g}\mathbb E_e
    \left\|y_e-P_g^\downarrow B_e h_e\right\|_2^2,
    \quad (P_g^\downarrow)^\top P_g^\downarrow=I_{r_g}.
\end{equation}
Here $B_e\in\mathbb R^{r_g\times\dff}$ denotes the temporary projections before intermediate-width compression.
The joint down problem admits a closed-form solution.
Eliminating $B_e$ reduces Obj.~\ref{obj:down_objective} to finding the leading eigenvectors of the summed output second moments.
The optimal shared and temporary expert-specific projections are
\begin{equation}
    \label{eq:down_solution_main}
    P_g^\downarrow=\operatorname{eig}_{r_g}\!\left(
    \sum_{e\in\mathcal E_g}C^y_e\right),
    \qquad B_e=(P_g^\downarrow)^\top W_e^\downarrow.
\end{equation}
This solution minimizes the objective before width compression.
Appendix~\ref{app:down_solution} derives this solution.

Intermediate-width compression introduces a further approximation: the full hidden activation is reconstructed from the selected coordinates using Nystr\"om approximation \citep{bib:NystromMatrix}, giving the final expert-specific projections:
\begin{equation}
    \label{eq:nystrom}
    A_e^\downarrow=B_e C^h_eS_e(S_e^\top C^h_eS_e)^\dagger
    \in\mathbb R^{r_g\times k_e},
\end{equation}
where $\dagger$ denotes the Moore--Penrose pseudoinverse; the reconstruction is derived in Appendix~\ref{app:width_solution}.
Thus, down-width reconstruction follows fitting the shared output projection, while up/gate selection precedes factorization.
All statistics come from the original model.

\subsection{Error-Guided Expert Grouping}
\label{sec:grouping}

We assign experts to the $G$ groups by minimizing reconstruction error.
For fixed neuron selections and candidate shared projections, $A_e^\uparrow$ and $B_e$ take their optimal values for each candidate group.
Writing $g=g(e)$ within each summand, the assignment satisfies
\begin{equation}
    \label{obj:grouping_objective}
    \begin{aligned}
            & \operatorname*{arg\,min}_{g(\cdot)}
        \sum_e\mathbb E_e\!\left[
            \|S_e^\top W_e^\uparrow x-A_e^\uparrow(P_g^\uparrow)^\top x\|_2^2
            +\|y_e-P_g^\downarrow B_eh_e\|_2^2
        \right]                                   \\
        ={} & \operatorname*{arg\,max}_{g(\cdot)}
        \sum_e\underbrace{\left\{
            \begin{aligned}
                 & \tr\!\Big[
                    \left((P_g^\uparrow)^\top C^x_eP_g^\uparrow\right)^{-1}
                    (P_g^\uparrow)^\top C^x_e(W_e^\uparrow)^\top
                    S_eS_e^\top W_e^\uparrow C^x_eP_g^\uparrow
                \Big]                                                           \\
                 & +\tr\!\left[(P_g^\downarrow)^\top C^y_eP_g^\downarrow\right]
            \end{aligned}
            \right\}}_{J_e^g}.
    \end{aligned}
\end{equation}
Here $J_e^g$ is the reconstruction-error reduction after fitting $A_e^\uparrow$ and $B_e$, and the assignment satisfies the group-size constraints.
The gate projection contributes an analogous term.
Appendix~\ref{app:grouping_derivation} provides the derivation.
We use this error-reduction criterion to construct the pairwise assignment scores below, then fit shared projections to the resulting groups using Section~\ref{sec:factorization}.

\textbf{Up/down balance.}
Neuron selection reduces the up term of Obj.~\ref{obj:grouping_objective}.
To balance the grouping scores, we multiply their down term by the retained pre-activation second-moment fraction of the up projections:
\begin{equation}
    \label{eq:rho}
    \beta_{\mathrm{scale}}=
    \frac{\sum_{e=1}^E
        \tr\!\left(S_e^\top W_e^\uparrow C_e^x(W_e^\uparrow)^\top S_e\right)}
    {\sum_{e=1}^E
        \tr\!\left(W_e^\uparrow C_e^x(W_e^\uparrow)^\top\right)}.
\end{equation}
For gated experts, we add the corresponding gate term to both the numerator and denominator, using the same neuron selections.
This gives the error-reduction score for a reconstruction objective whose down error is weighted by $\beta_{\mathrm{scale}}$.
The scaling is used only for grouping; projection fitting and rank allocation retain the original $C_e^y$.

\textbf{Group assignment.}
For experts $i,j\in\{1,\dots,E\}$, let $J_i^j$ be the score from Obj.~\ref{obj:grouping_objective}, with the up/down balancing above, for expert $i$ evaluated on the rank-$r$ subspace fitted to expert $j$.
We define the pairwise dissimilarity matrix $D\in\R^{E\times E}$ by $D_{i,j}=1-J_i^j/J_i^i$, where $J_i^i$ is the self-reconstruction score.
Because the up subspace is fitted approximately, this normalization does not guarantee that $D_{i,j}$ lies in $[0,1]$.
Groups are then obtained by balanced medoid clustering on $D$ with cluster sizes fixed to $E/G$ and assignment solved exactly by the Jonker-Volgenant \citep{bib:JonkerVolgenant} algorithm over medoids replicated into $E / G$ slots.
We initialize medoids using $k$-means++ seeding and alternate balanced assignment with medoid updates. The pairwise scores require $E$ single-expert subspace fits and an $E\times E$ gain sweep.

\subsection{Capacity Allocation}
\label{sec:allocation}

\textbf{Within a layer.}
Assume each expert has $n$ weight matrices, each containing $d\cdot \dff$ parameters, with dimensions $\dff\times d$ or $d\times\dff$.
Each matrix has its own shared and expert-specific projections, while all matrices use the same group rank $r_g$ and retained width $k_e$.
The compressed weights therefore contain $n\sum_g r_gc_g$ parameters, where $c_g=d+\sum_{e\in\mathcal E_g}k_e$.

For fixed groups and neuron selections, rank allocation uses the matrices underlying the projection solutions:
\begin{equation}
    \label{eq:rank_matrices}
    \begin{aligned}
        H_g^\uparrow   & =C_g^{1/2}\left(\sum_{e\in\mathcal E_g}\tau_e
        (W_e^\uparrow)^\top S_eS_e^\top W_e^\uparrow\right)C_g^{1/2},  \\
        H_g^\downarrow & =\sum_{e\in\mathcal E_g}C^y_e.
    \end{aligned}
\end{equation}
Here $C_g$ and $\tau_e$ are defined in Section~\ref{sec:factorization}.
The leading eigenvalues of $H_g^\uparrow$ measure the reduction in the up surrogate, whereas those of $H_g^\downarrow$ measure the reduction in Obj.~\ref{obj:down_objective} before width compression.
For matrix $j\in\{1,\ldots,n\}$, let $H_g^{(j)}$ denote the corresponding construction using that matrix's weights and calibration statistics.
We define $V_g(t)=\sum_{j=1}^n\sum_{i=1}^t\lambda_i(H_g^{(j)})$, with eigenvalues in descending order.
This is a rank-allocation score, not the exact error reduction $\sum_e J_e^g$ in Section~\ref{sec:grouping}, since its up terms use the surrogate.
Under the uniform-rank reference budget, we solve
\begin{equation}
    \label{obj:rank_alloc}
    \begin{aligned}
        \max_{\{r_g\}}\quad &
        \sum_g V_g(r_g)       \\
        \text{s.t.}\quad    &
        \sum_g r_gc_g\le r\sum_g c_g,\qquad
        r_{\min}\le r_g\le r_{\max},\quad r_g\in\mathbb Z.
    \end{aligned}
\end{equation}
The common factor $n$ cancels from the budget constraint.
Widths are selected first using the neuron scores under the total budget $K$, fixing both $c_g$ and the selections used in $H_g^{(j)}$ during rank allocation.
We approximately solve Obj.~\ref{obj:rank_alloc} by bisecting a Lagrange multiplier for the parameter budget, following the multiplier-search approach to resource allocation \citep{bib:Patriksson2008}, then greedily allocate the remaining budget. Appendix~\ref{app:rank_solver} describes the discrete allocation steps.

\textbf{Across layers.}
\label{layer_importance_scoring}
We now introduce the layer index $\ell$.
Using the gate-weighted expert output criterion of REAP \citep{bib:REAP}, normalized by input magnitude, we define
\begin{equation}
    \label{eq:layer_score}
    I_\ell=\sum_{e=1}^{E_\ell}\mathbb E_{\ell,e}
    \left[\frac{\pi_{\ell,e}(x)\|f_{\ell,e}(x)\|_2}{\|x\|_2}\right],
\end{equation}
where $\pi_{\ell,e}(x)$ is the routing weight and $\mathbb E_{\ell,e}$ averages over inputs routed to that expert.
Given a target parameter budget, we first choose the overall width retention and determine the corresponding rank budget, accounting for the shared projections.
We then allocate rank and width across layers using $I_\ell$, assigning less compression to more important layers.
Because parameter count depends jointly on rank and width, we keep the width allocation fixed and rescale the layer ranks to match the total budget in the continuous relaxation.
The resulting $r_\ell$ and $K_\ell$ guide the within-layer allocation of group ranks and expert widths.
Integer rounding and unused allocation slack can change the achieved compression.
Appendix~\ref{app:layer_allocation} gives the allocation and budget-correction details.

%% file: Sections/Results.tex
\section{Experiments}
\label{sec:experiments}
\input{Sections/Tables/MainResults}

\subsection{Settings}
\label{sec:settings}

\textbf{Models.}
We evaluate GS-MoE on Mixtral-8x7B, DeepSeek-MoE-16B, and Qwen2-57B-A14B, covering both coarse-grained and fine-grained MoE architectures.
We evaluate compression along shared-subspace rank and intermediate width separately in Table ~\ref{tab:factorization_nystrom} and jointly in Table ~\ref{tab:main_results}.

\input{Sections/Tables/GroupedAssignment}
\textbf{Implementations.}
Unless otherwise specified, compression and evaluation use one to two NVIDIA B200 GPUs, with compression requiring a single GPU.
For neuron selection in Section~\ref{sec:factorization}, we retain at least $5\%$ of each expert's intermediate width and choose $\eta\in\{0.01,0.1\}$.
The balanced medoid clustering in Section~\ref{sec:grouping} uses $8$ restarts and at most $25$ Lloyd iterations, with pairwise gains computed in FP32.
For rank allocation in Section~\ref{sec:allocation}, group ranks range from $\max(\lfloor0.35\cdot r_\ell\rfloor,8)$ to $\min(\lfloor2.5\cdot r_\ell\rfloor,d)$.
GS-MoE requires no post-compression training, and we report achieved compression after rounding and discrete allocation.
Additional numerical and allocation settings are provided in Appendix~\ref{app:experimental_settings}.
Reported compression rates measure the reduction in total model parameters relative to the uncompressed model.

\textbf{Datasets.}
We collect calibration statistics in FP64 using a single forward pass over $512$ Alpaca samples \citep{bib:Alpaca}, each containing $2048$ tokens.
We evaluate on nine datasets, including C4 \citep{bib:C4-dataset} and WikiText-2-raw-v1\citep{bib:Wikitext} for language modeling, and use the LM Evaluation Harness \citep{bib:LM-Eval} for zero-shot tasks OpenbookQa \citep{bib:OpenbookQA}, HellaSwag \citep{bib:HellaSwag}, ARC-c \citep{bib:ArcChallenge}, ARC-e, PIQA \citep{bib:PIQA}, Winogrande \citep{bib:WinoGrande}, and mathqa \citep{bib:MathQA}.

\textbf{Baselines.}
We compare GS-MoE with MoE-SVD \citep{bib:MoE-SVD}, D$^2$-MoE \citep{bib:D2-MoE}, TD-MoE \citep{bib:TD-MoE}, NAEE \citep{bib:NAEE}, and MoE-I$^2$ \citep{bib:MoE-I2}.
\input{Sections/TexFigures/MainFigure}

\subsection{Results}
\textbf{Zero-shot performance.} We evaluated GS-MoE on seven zero-shot reasoning tasks and compared it to leading MoE compression baselines.
Table \ref{tab:main_results} demonstrates that GS-MoE achieves superior accuracy at higher compression ratios on Qwen2-57B-A14B and Deepseek-MoE-16B-Base, with Qwen2 attaining lossless accuracy at 60\% compression and Deepseek retaining 90\% of reasoning accuracy.
We compare Deepseek-MoE-16B-Base and Qwen2-57B-A14B to $D^2$-MoE, which utilizes a single base-weight plus expert-specific residuals, and find that GS-MoE outperforms $D^2$-MoE significantly at all compression ratios for these models.
GS-MoE effectively compresses Mixtral-8x7B at compression ratios up to 40\%, matching or beating other leading methods.

\textbf{Sensitivity to clustering iterations and restarts.}
We evaluate the number of Lloyd-style iterations~\citep{bib:Lloyd1982} and independent restarts with $k$-means++ seeding~\citep{bib:Arthur2007} in our balanced medoid clustering.
As shown in Figure~\ref{fig:iters_and_restarts}, two iterations with two restarts recover 98\% of the largest gain observed across the tested configurations.
The lowest perplexity is obtained with four iterations and eight restarts; further iterations slightly increase perplexity.

\begin{figure*}[t]
  \centering
  \begin{subfigure}[t]{0.48\linewidth}
    \centering
    \includegraphics[width=\linewidth]{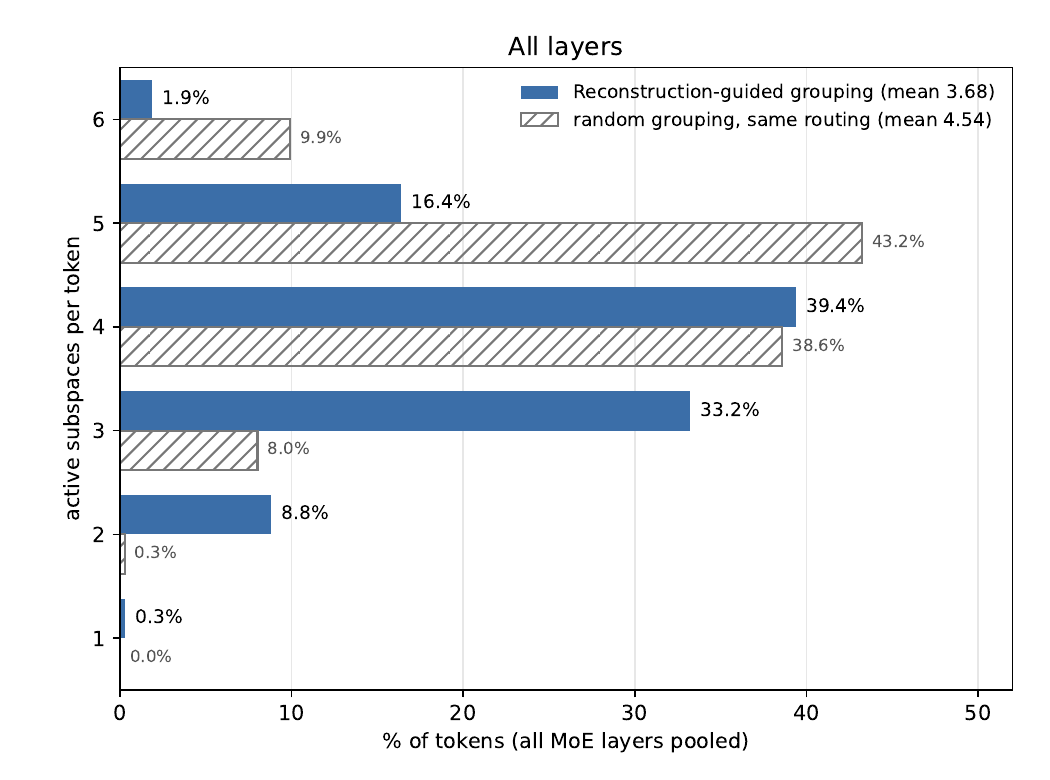}
    \caption{Active subspaces per token.}
    \label{fig:active_subspace_p_tok}
  \end{subfigure}
  \hfill
  \begin{subfigure}[t]{0.48\linewidth}
    \centering
    \includegraphics[width=\linewidth]{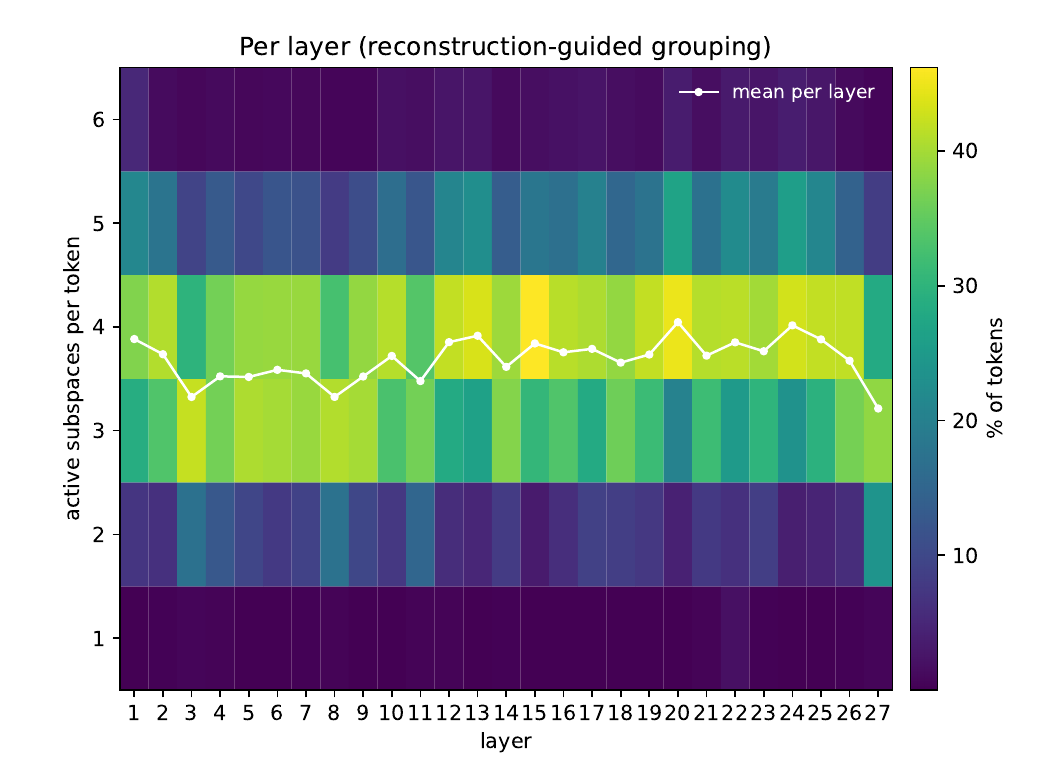}
    \caption{Active subspaces across all MoE Layers}
    \label{fig:active_subspaces_all_layers}
  \end{subfigure}
  \caption{Active subspaces per token on DeepSeek-MoE-16B-Base ($G=8$), counted as distinct groups among the top-6 experts.
    (a) Guided versus random grouping over all tokens and layers.
    (b) Per-layer distributions for guided grouping, with means shown in the white line.}
  \label{fig:active_subspaces_tok_breakdown}
  \vspace{-1em}
\end{figure*}

\input{Sections/Tables/Throughput}

\textbf{Grouped assignment and subspace rank allocation.}
Table~\ref{tab:ablation_grouped_rank} shows that grouped assignment reduces WikiText perplexity from 29.49 to 25.23.
Adding group rank allocation further improves perplexity across all three datasets, with an average reduction of 0.57.
Without grouped assignment, rank allocation provides no consistent improvement.

\textbf{Runtime efficiency analysis.}
Table~\ref{tab:throughput} reports throughput at 60\% compression. For each model, we benchmark all methods under identical inference settings with B200 GPU(s) for Mixtral-8x7B and H200 GPU(s) for DeepSeek-MoE-16B-Base.
On Mixtral-8x7B, GS-MoE achieves 374.77 tokens/sec with $G=1$, a $3.32\times$ speedup over D$^2$-MoE (112.84) and $3.66\times$ over TD-MoE (102.53), and 6.6\% higher than MoE-SVD (351.44); increasing $G$ to 2 reduces throughput by 2.9\%.
On DeepSeek-MoE-16B-Base, increasing $G$ from 1 to 8 reduces throughput by only 9.6\%.
Figure~\ref{fig:active_subspace_p_tok} suggests a possible explanation for this moderate overhead: with $G=8$, each token activates an average of only 3.68 subspaces.
Our grouping reduces this average by approximately 19\% relative to random grouping under identical routing.

%% file: Sections/Tables/MainResults.tex
\begin{table*}[!t]
\caption{Comparison of GS-MoE to other methods. Best results are in \textbf{bold}.}
\label{tab:main_results}
\vspace{3pt}
\centering
\scriptsize
\setlength{\tabcolsep}{8pt}
\renewcommand{\arraystretch}{1.125}  %
\setlength{\arrayrulewidth}{0.4pt}
\setlength{\dashlinedash}{2pt}  %
\setlength{\dashlinegap}{2pt}   %
\def\rulegap{1pt}  %
\def\dashgap{1pt}    %
\def\ruletop{\vrule width 0pt height \dimexpr\arraystretch\ht\strutbox+\rulegap\relax depth 0pt\relax}  %
\def\rulebot{\vrule width 0pt height 0pt depth \dimexpr\arraystretch\dp\strutbox+\rulegap\relax\relax}  %
\def\dashtop{\vrule width 0pt height \dimexpr\arraystretch\ht\strutbox+\dashgap\relax depth 0pt\relax}  %
\def\dashbot{\vrule width 0pt height 0pt depth \dimexpr\arraystretch\dp\strutbox+\dashgap\relax\relax}  %
\resizebox{\ifdim\width>\linewidth\linewidth\else\width\fi}{!}{%
\begin{tabular}{c|l|rr|rrrrrrrr}
\noalign{\global\arrayrulewidth=0.8pt}\hline\noalign{\global\arrayrulewidth=0.4pt}
\ruletop\rulebot Ratio & Method & Wiki.$\downarrow$ & PTB$\downarrow$ & OpenB. & ARC-e & WinoG. & HellaS. & ARC-c & PIQA & MathQA & Avg.$\uparrow$ \\
\hline
\multicolumn{12}{c}{\ruletop\rulebot\textbf{DeepSeekMoE-16B-Base} (64 experts, top-6 activated)} \\
\hline
\ruletop\rulebot 0\% & Original & 6.38 & 9.47 & 0.32 & 0.75 & 0.70 & 0.58 & 0.44 & 0.79 & 0.31 & 0.56 \\
\hline
\ruletop\multirow{4}{*}[-\dashgap]{20\%}
 & NAEE~\citeyearpar{bib:NAEE} & 9.44 & 15.02 & 0.32 & 0.71 & 0.66 & 0.55 & 0.40 & 0.77 & 0.29 & 0.53 \\
 & MoE-I$^2$~\citeyearpar{bib:MoE-I2} & 7.69 & 11.59 & 0.26 & 0.71 & 0.68 & 0.49 & 0.38 & 0.73 & 0.29 & 0.51 \\
\dashbot & D$^2$-MoE~\citeyearpar{bib:D2-MoE} & \textbf{6.84} & 11.10 & 0.30 & 0.74 & 0.69 & 0.55 & 0.41 & 0.76 & \textbf{0.31} & 0.54 \\
\cdashline{2-12}
\dashtop\rulebot & \textbf{GS-MoE (Ours)} & 7.35 & \textbf{10.64} & \textbf{0.33} & \textbf{0.75} & \textbf{0.70} & \textbf{0.57} & \textbf{0.43} & \textbf{0.79} & 0.30 & \textbf{0.55} \\
\hline
\ruletop\multirow{4}{*}[-\dashgap]{40\%}
 & NAEE~\citeyearpar{bib:NAEE} & 8.55 & 14.47 & 0.23 & 0.67 & 0.67 & 0.41 & 0.32 & 0.69 & 0.26 & 0.46 \\
 & MoE-I$^2$~\citeyearpar{bib:MoE-I2} & 9.73 & 15.75 & 0.23 & 0.64 & 0.66 & 0.41 & 0.31 & 0.68 & 0.26 & 0.46 \\
\dashbot & D$^2$-MoE~\citeyearpar{bib:D2-MoE} & \textbf{7.93} & 14.07 & 0.26 & 0.69 & 0.65 & 0.45 & 0.36 & 0.72 & 0.28 & 0.49 \\
\cdashline{2-12}
\dashtop\rulebot & \textbf{GS-MoE (Ours)} & 10.10 & \textbf{13.51} & \textbf{0.31} & \textbf{0.75} & \textbf{0.70} & \textbf{0.53} & \textbf{0.43} & \textbf{0.78} & \textbf{0.29} & \textbf{0.54} \\ %
\hline
\ruletop\multirow{4}{*}[-\dashgap]{60\%}
 & NAEE~\citeyearpar{bib:NAEE} & 23.20 & 49.89 & 0.17 & 0.49 & 0.58 & 0.33 & 0.24 & 0.61 & 0.23 & 0.38 \\
 & MoE-I$^2$~\citeyearpar{bib:MoE-I2} & 15.83 & 32.20 & 0.17 & 0.48 & 0.58 & 0.32 & 0.23 & 0.61 & 0.22 & 0.37 \\
\dashbot & D$^2$-MoE~\citeyearpar{bib:D2-MoE} & \textbf{11.67} & 27.73 & 0.21 & 0.54 & 0.61 & 0.35 & 0.29 & 0.63 & 0.24 & 0.41 \\
\cdashline{2-12}
\dashtop\rulebot & \textbf{GS-MoE (Ours)} & 17.55 & \textbf{22.21} & \textbf{0.28} & \textbf{0.72} & \textbf{0.66} & \textbf{0.45} & \textbf{0.39} & \textbf{0.74} & \textbf{0.28} & \textbf{0.50} \\
\hline
\multicolumn{12}{c}{\ruletop\rulebot\textbf{Qwen2-57B-A14B} (64 experts, top-8 activated)} \\
\hline
\ruletop\rulebot 0\% & Original & 5.12 & 9.18 & 0.33 & 0.75 & 0.74 & 0.63 & 0.46 & 0.81 & 0.39 & 0.59 \\
\hline
\ruletop\multirow{4}{*}[-\dashgap]{40\%}
 & NAEE~\citeyearpar{bib:NAEE} & 8.79 & 12.64 & 0.28 & 0.72 & 0.71 & 0.54 & 0.39 & 0.76 & 0.32 & 0.53 \\
 & TD-MoE~\citeyearpar{bib:TD-MoE} & 8.07 & 12.50 & 0.29 & \textbf{0.77} & 0.71 & 0.54 & 0.45 & 0.78 & 0.36 & 0.56 \\
\dashbot & D$^2$-MoE~\citeyearpar{bib:D2-MoE} & 8.19 & 11.23 & \textbf{0.34} & 0.75 & \textbf{0.75} & 0.61 & 0.45 & 0.79 & 0.36 & 0.58 \\
\cdashline{2-12}
\dashtop\rulebot & \textbf{GS-MoE (Ours)} & \textbf{6.59} & \textbf{9.81} & \textbf{0.34} & \textbf{0.77} & \textbf{0.75} & \textbf{0.62} & \textbf{0.47} & \textbf{0.81} & \textbf{0.39} & \textbf{0.59} \\
\hline
\ruletop\multirow{3}{*}[-\dashgap]{60\%}
 & NAEE~\citeyearpar{bib:NAEE} & 14.79 & 23.52 & 0.21 & 0.58 & 0.61 & 0.44 & 0.29 & 0.69 & 0.26 & 0.44 \\
\dashbot & TD-MoE~\citeyearpar{bib:TD-MoE} & 12.49 & 19.92 & 0.28 & 0.73 & 0.68 & 0.46 & 0.41 & 0.74 & 0.31 & 0.52 \\
\cdashline{2-12}
\dashtop\rulebot & \textbf{GS-MoE (Ours)} & \textbf{7.87} & \textbf{11.29} & \textbf{0.32} & \textbf{0.79} & \textbf{0.78} & \textbf{0.59} & \textbf{0.50} & \textbf{0.80} & \textbf{0.38} & \textbf{0.59} \\ %
\hline
\multicolumn{12}{c}{\ruletop\rulebot\textbf{Mixtral-8$\times$7B-v0.1} (8 experts, top-2 activated)} \\
\hline
\ruletop\rulebot 0\% & Original & 3.98 & 12.99 & 0.36 & 0.84 & 0.76 & 0.65 & 0.57 & 0.82 & 0.43 & 0.63 \\
\hline
\ruletop\multirow{4}{*}[-\dashgap]{20\%}
 & MoE-SVD~\citeyearpar{bib:MoE-SVD} & 4.86 & 19.42 & 0.33 & 0.79 & 0.74 & 0.56 & 0.49 & 0.78 & 0.37 & 0.58 \\
 & TD-MoE~\citeyearpar{bib:TD-MoE} & \textbf{4.49} & 17.20 & 0.33 & \textbf{0.83} & \textbf{0.77} & \textbf{0.64} & \textbf{0.53} & \textbf{0.82} & 0.40 & \textbf{0.62} \\
\dashbot & D$^2$-MoE~\citeyearpar{bib:D2-MoE} & 4.65 & 16.32 & 0.33 & 0.80 & 0.75 & 0.61 & 0.51 & 0.81 & 0.39 & 0.60 \\
\cdashline{2-12}
\dashtop\rulebot & \textbf{GS-MoE (Ours)} & 5.88 & \textbf{15.69} & \textbf{0.36} & 0.82 & 0.76 & 0.62 & \textbf{0.53} & \textbf{0.82} & \textbf{0.41} & \textbf{0.62} \\
\hline
\ruletop\multirow{4}{*}[-\dashgap]{40\%}
 & MoE-SVD~\citeyearpar{bib:MoE-SVD} & 6.74 & 27.73 & 0.27 & 0.72 & 0.67 & 0.43 & 0.38 & 0.71 & 0.32 & 0.50 \\
 & TD-MoE~\citeyearpar{bib:TD-MoE} & 5.79 & 24.60 & 0.28 & 0.77 & \textbf{0.76} & \textbf{0.57} & 0.47 & \textbf{0.79} & 0.35 & \textbf{0.57} \\
\dashbot & D$^2$-MoE~\citeyearpar{bib:D2-MoE} & \textbf{5.28} & \textbf{20.54} & \textbf{0.32} & 0.78 & 0.73 & \textbf{0.57} & 0.47 & 0.78 & 0.34 & \textbf{0.57} \\
\cdashline{2-12}
\dashtop\rulebot & \textbf{GS-MoE (Ours)} & 9.29 & 22.17 & 0.31 & \textbf{0.79} & 0.70 & 0.55 & \textbf{0.49} & \textbf{0.79} & \textbf{0.38} & \textbf{0.57} \\
\hline
\ruletop\multirow{4}{*}[-\dashgap]{60\%}
 & MoE-SVD~\citeyearpar{bib:MoE-SVD} & 13.52 & 130.26 & 0.19 & 0.45 & 0.55 & 0.33 & 0.23 & 0.62 & 0.25 & 0.37 \\
 & TD-MoE~\citeyearpar{bib:TD-MoE} & 17.78 & 79.43 & 0.21 & 0.58 & 0.62 & 0.38 & 0.28 & 0.65 & 0.27 & 0.43 \\
\dashbot & D$^2$-MoE~\citeyearpar{bib:D2-MoE} & \textbf{6.46} & \textbf{23.63} & \textbf{0.28} & \textbf{0.72} & \textbf{0.71} & \textbf{0.51} & 0.38 & \textbf{0.73} & 0.31 & \textbf{0.52} \\
\cdashline{2-12}
\dashtop\rulebot & \textbf{GS-MoE (Ours)} & 15.94 & 35.96 & 0.26 & \textbf{0.72} & 0.64 & 0.44 & \textbf{0.41} & \textbf{0.73} & \textbf{0.32} & 0.50 \\
\noalign{\global\arrayrulewidth=0.8pt}\hline\noalign{\global\arrayrulewidth=0.4pt}
\end{tabular}%
}
\vspace{-10pt}
\end{table*}

%% file: Sections/Tables/GroupedAssignment.tex
\begin{wraptable}{r}{0.5\textwidth}
\centering
\caption{Ablation of grouped assignment and group rank allocation
for GS-MoE factorization only on DeepSeek-MoE-16B-Base
at 60\% compression with $G=8$. Lower perplexity is better.}
\label{tab:ablation_grouped_rank}
\small
\setlength{\tabcolsep}{4pt}
\renewcommand{\arraystretch}{1.15}
\begin{tabular}{ccccc}
\toprule
\shortstack{Grouped\\assignment} &
\shortstack{Group rank\\allocation} &
Wiki. & C4 & PTB \\
\midrule
\ding{55} & \ding{55} & 29.49 & 33.64 & 36.49 \\
\ding{55} & \ding{51} & 29.42 & 33.89 & 36.56 \\
\ding{51} & \ding{55} & 25.23 & 28.83 & 30.35 \\
\ding{51} & \ding{51} &
\textbf{25.02} & \textbf{28.30} & \textbf{29.37} \\
\bottomrule
\end{tabular}
\end{wraptable}

%% file: Sections/TexFigures/MainFigure.tex
\begin{figure*}[t]
  \centering
  \begin{subfigure}[t]{0.31\linewidth}
    \centering
    \includegraphics[width=\linewidth]{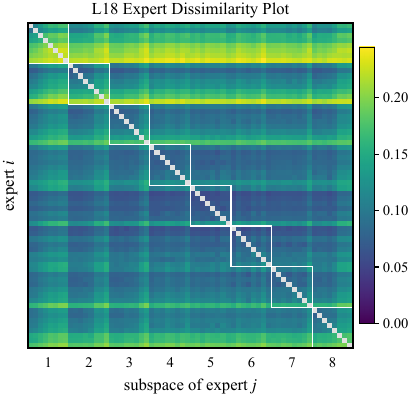}
    \caption{Qwen2, layer 18}
    \label{fig:dissim_qwen_l18}
  \end{subfigure}
  \hfill
  \begin{subfigure}[t]{0.31\linewidth}
    \centering
    \includegraphics[width=\linewidth]{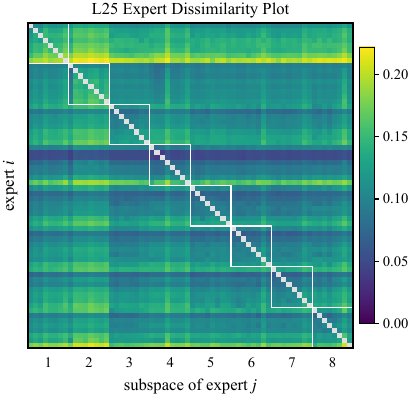}
    \caption{Qwen2, layer 25}
    \label{fig:dissim_qwen_l25}
  \end{subfigure}
  \hfill
  \begin{subfigure}[t]{0.33\linewidth}
    \centering
    \includegraphics[width=\linewidth]{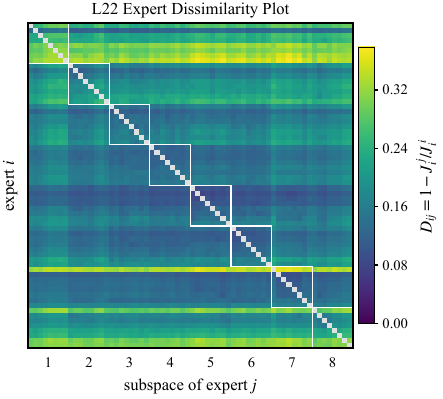}
    \caption{DeepSeek, layer 22}
    \label{fig:dissim_deepseek_l22}
  \end{subfigure}
  \vspace{0.8em}
  \hspace{-2.25em}
  \begin{subfigure}[t]{0.31\linewidth}
    \centering
    \includegraphics[height=4.6cm]{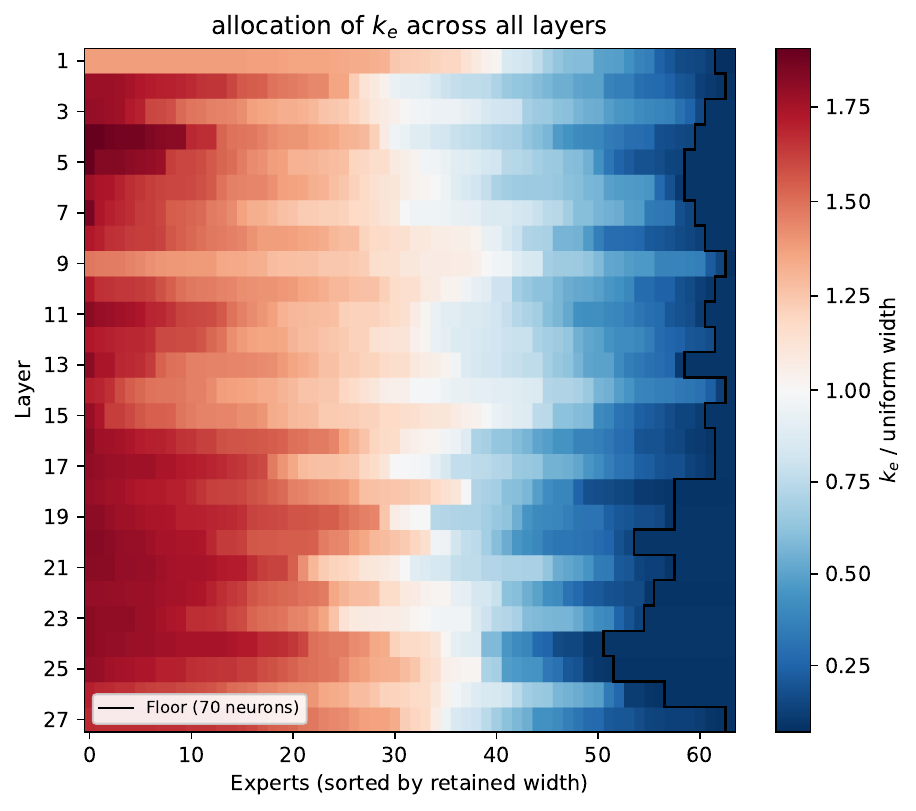}
    \caption{Width allocation per layer}
    \label{fig:width_heatmap}
  \end{subfigure}
  \hspace{0.4em}
  \begin{subfigure}[t]{0.31\linewidth}
    \centering
    \includegraphics[height=4.6cm]{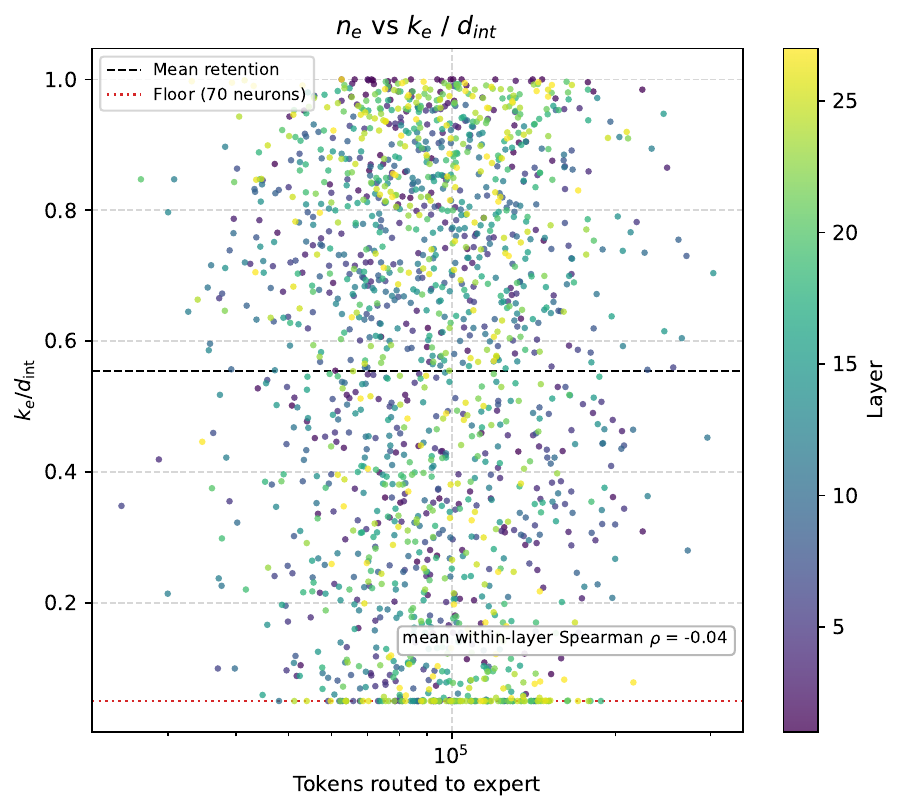}
    \caption{Width vs.\ routing load}
    \label{fig:width_scatter}
  \end{subfigure}
  \begin{subfigure}[t]{0.31\linewidth}
    \centering
    \includegraphics[height=4.6cm]{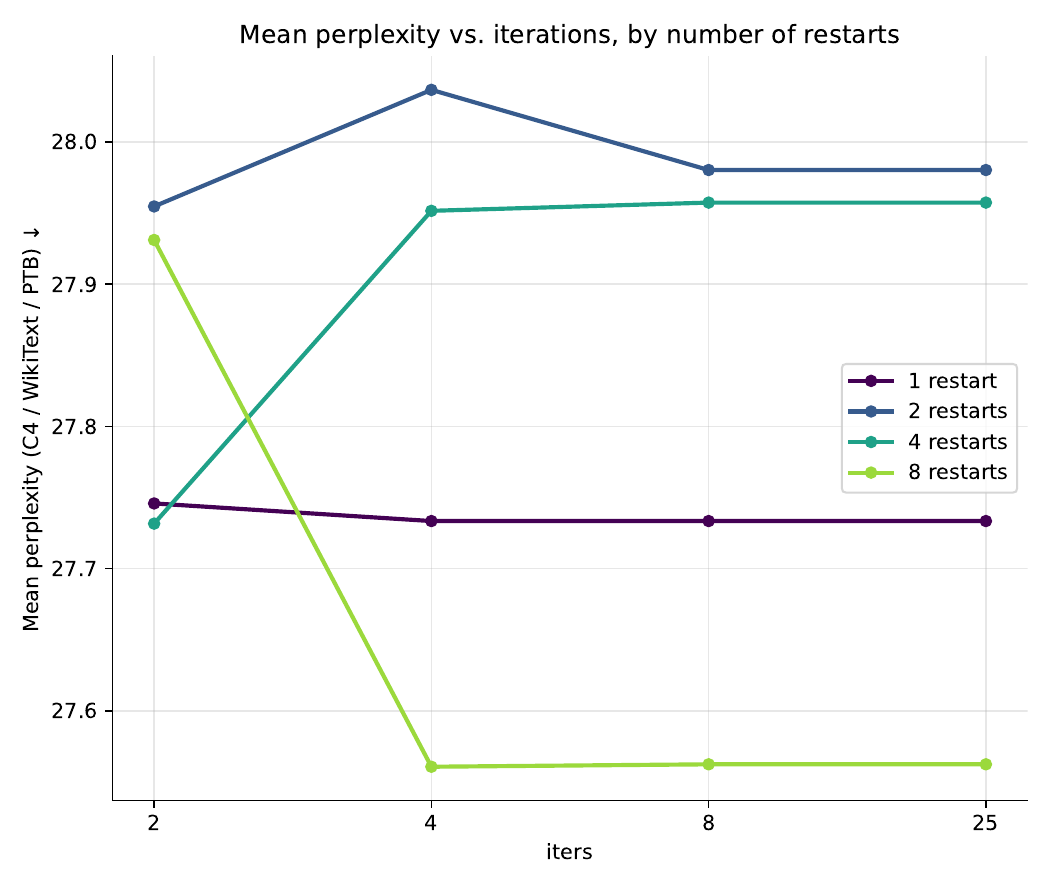}
    \captionsetup{margin={1.0em,0pt}} 
    \caption{Expert-subspace grouping}
  \label{fig:iters_and_restarts}   
  \end{subfigure}
  \caption{(a--c) Expert-to-expert dissimilarity via Obj.~\ref{obj:grouping_objective} on Qwen2-57B-A14B and DeepSeek-MoE-16B-Base.
White boxes indicate assigned groups; darker colors indicate lower dissimilarity.
(d) Leverage-based retained width $k_e$ relative to a uniform split within each layer on DeepSeek-MoE-16B-Base at 60\% compression ($\nu=0.55$), with experts sorted by width; the black line marks the 70-neuron floor.
(e) Under the same setting, $k_e/\dff$ versus tokens routed to each expert (mean within-layer Spearman $\rho=-0.04$).
(f) Sensitivity of expert-subspace grouping to the number of restarts and iterations.}
  \label{fig:expert_dissimilarity}
\end{figure*}

%% file: Sections/Tables/Throughput.tex
\begin{wraptable}{r}{0.40\textwidth}
    \centering
    \caption{Throughput (tokens/sec, $\uparrow$) at 60\% compression.}
    \label{tab:throughput}
    \footnotesize
    \setlength{\tabcolsep}{5pt}
    \setlength{\dashlinedash}{2pt}  %
    \setlength{\dashlinegap}{2pt}   %
    \def\dashbot{\vrule width 0pt height 0pt depth \dimexpr\arraystretch\dp\strutbox+\aboverulesep\relax\relax}  %
    \def\dashtop{\vrule width 0pt height \dimexpr\arraystretch\ht\strutbox+\belowrulesep+\arrayrulewidth\relax depth 0pt\relax}  %
    \begin{tabular}{@{}lr@{}}
        \toprule
        \multicolumn{2}{c}{\textbf{Mixtral-8x7B}}         \\
        \midrule
        Method                               & Tokens/sec \\
        \midrule
        D$^2$-MoE~\citep{bib:D2-MoE}         & 112.84     \\
        TD-MoE~\citep{bib:TD-MoE}            & 102.53     \\
        \dashbot MoE-SVD~\citep{bib:MoE-SVD} & 351.44     \\
        \cdashline{1-2}
        \dashtop
        \textbf{GS-MoE} ($G=1$)              & 374.77     \\
        \textbf{GS-MoE} ($G=2$)              & 363.78     \\
        \midrule
        \multicolumn{2}{c}{\textbf{DeepSeek-MoE-16B-Base}} \\
        \midrule
        GS-MoE ($G=1$)                       & 250.99     \\
        GS-MoE ($G=2$)                       & 239.79     \\
        GS-MoE ($G=4$)                       & 237.37     \\
        GS-MoE ($G=8$)                       & 226.81     \\
        \bottomrule
    \end{tabular}
\end{wraptable}

%% file: Sections/Conclusion.tex
\section{Conclusion}
In this work, we introduce GS-MoE, a novel training-free compression framework that addresses MoE parameter overhead through group-wise shared subspaces and intra-expert intermediate width reduction.
We propose a novel grouping framework that accurately groups experts into error-minimized shared subspaces.
Explicit group assignment drastically improves the efficiency of weight-sharing against fine-grained architectures and yields minimal throughput-overhead relative to accuracy improvement.
In addition to our expert-grouping framework, we provide a complementary eigenvalue-based group-rank allocation scheme and guide expert-width allocation through leverage scores.
We find that, for fine-grained architectures, combining group-wise subspaces and expert intermediate width reduction into a single compression framework retains state-of-the-art fidelity under aggressive parameter budgets.

%% file: Sections/Appendix.tex
\section{Method Details and Derivations}
\label{app:method_details}
\label{app:projection_solutions}

We use the notation of Section~\ref{sec:factorization} and fix one group $\mathcal E_g$, its rank $r_g$, and the neuron selections $S_e$.
The statistics $C^x_e$, $C^h_e$, and $C^y_e$ are computed from the original model.
The final expert-specific projections have dimensions $A_e^\uparrow\in\mathbb R^{k_e\times r_g}$ and $A_e^\downarrow\in\mathbb R^{r_g\times k_e}$; only the temporary down projections $B_e\in\mathbb R^{r_g\times\dff}$ retain the original width.
We write $\operatorname{eig}_r(H)$ for the leading $r$ orthonormal eigenvectors of a symmetric matrix, and $\operatorname{orth}(M)$ for an orthonormal basis of its column space.
Inverse expressions below assume that the relevant matrices are nonsingular.

\subsection{Up Projection Solution}
\label{app:up_solution}

\paragraph{Expert-specific projections for a fixed shared projection.}
Differentiating the up reconstruction error with respect to $A_e^\uparrow$ yields the normal equations
$A_e^\uparrow((P_g^\uparrow)^\top C^x_eP_g^\uparrow)
    =S_e^\top W_e^\uparrow C^x_eP_g^\uparrow$.
Consequently,
\begin{equation}
    \label{eq:Aup}
    A_e^\uparrow=S_e^\top W_e^\uparrow C^x_eP_g^\uparrow
    \left((P_g^\uparrow)^\top C^x_eP_g^\uparrow\right)^{-1}.
\end{equation}
The least-squares problem is convex in the expert-specific projections, so this solution minimizes each expert's error for fixed $P_g^\uparrow$.
With singular projected statistics, the pseudoinverse gives a minimizing solution instead.

\paragraph{Why the shared projection requires an approximation.}
Substituting Eq.~(\ref{eq:Aup}) into Obj.~\ref{obj:up_objective} leaves a projection-independent term minus
\begin{equation}
    \label{eq:up_retained_energy}
    \sum_{e\in\mathcal E_g}\tr\!\left[
        \left((P_g^\uparrow)^\top C^x_eP_g^\uparrow\right)^{-1}
        (P_g^\uparrow)^\top C^x_e(W_e^\uparrow)^\top
        S_eS_e^\top W_e^\uparrow C^x_eP_g^\uparrow
        \right].
\end{equation}
The inverse term depends on both the expert and the shared projection.
For general heterogeneous $C^x_e$, this expression does not reduce to a single fixed-matrix eigenproblem.
We therefore use the group-level approximation below to estimate the shared projection.

\paragraph{Group-level surrogate.}
Let $\tau_e=n_e/\sum_{j\in\mathcal E_g}n_j$, where $n_e$ counts routed calibration assignments, and set $C_g=\sum_{e\in\mathcal E_g}\tau_e C^x_e$.
Replacing the expert input statistics by $C_g$ and weighting expert contributions by $\tau_e$ gives the surrogate
\begin{equation}
    \label{obj:up_surrogate}
    \min_{P_g^\uparrow,\{A_e^\uparrow\}}
    \sum_{e\in\mathcal E_g}\tau_e
    \left\|(S_e^\top W_e^\uparrow-A_e^\uparrow(P_g^\uparrow)^\top)
    C_g^{1/2}\right\|_F^2,
    \qquad (P_g^\uparrow)^\top P_g^\uparrow=I_{r_g}.
\end{equation}
Assume $C_g\succ0$.
For this derivation only, let $Q$ be an orthonormal basis of the column space of $C_g^{1/2}P_g^\uparrow$.
After minimizing over the expert-specific projections, the surrogate becomes
$\tr(H_g^\uparrow)-\tr(Q^\top H_g^\uparrow Q)$, where
\begin{equation}
    \label{eq:energy_up}
    H_g^\uparrow=C_g^{1/2}
    \left(\sum_{e\in\mathcal E_g}\tau_e
        (W_e^\uparrow)^\top S_eS_e^\top W_e^\uparrow\right)C_g^{1/2}.
\end{equation}
Its minimum is the eigenvalue tail $\sum_{i>r_g}\lambda_i(H_g^\uparrow)$, attained by the leading eigenspace.
Mapping this space back gives the shared projection in Eq.~(\ref{eq:up_solution}).
Orthonormalization changes the coordinates but not the represented subspace.
We then refit the expert-specific projections with the original $C^x_e$ using Eq.~(\ref{eq:Aup}).
This projection is optimal for the surrogate, not necessarily for the original up objective.
The same derivation applies to the gate projection with the same selected neurons.

\subsection{Down Projection Solution}
\label{app:down_solution}

\paragraph{Eliminating the temporary projections.}
For any orthonormal $P_g^\downarrow$, decompose the output error into orthogonal components:
\begin{equation}
    \label{eq:down_error_split}
    \begin{aligned}
        \mathbb E_e\|y_e-P_g^\downarrow B_eh_e\|_2^2
        ={} & \mathbb E_e\|(I-P_g^\downarrow(P_g^\downarrow)^\top)y_e\|_2^2 \\
            & +\mathbb E_e\|(P_g^\downarrow)^\top y_e-B_eh_e\|_2^2.
    \end{aligned}
\end{equation}
Since $y_e=W_e^\downarrow h_e$, choosing $B_e=(P_g^\downarrow)^\top W_e^\downarrow$ makes the second term zero.
This choice of $B_e$ globally minimizes the error for the fixed shared projection even when $C^h_e$ is singular; uniqueness is not required.

\paragraph{Optimal shared projection.}
The remaining unweighted objective is
$\sum_e\tr(C^y_e)-\tr((P_g^\downarrow)^\top(\sum_e C^y_e)P_g^\downarrow)$.
Hence a global solution to Obj.~\ref{obj:down_objective} is
\begin{equation}
    \label{eq:down_solution}
    P_g^\downarrow=\operatorname{eig}_{r_g}\!\left(
    \sum_{e\in\mathcal E_g}C^y_e\right),
    \qquad B_e=(P_g^\downarrow)^\top W_e^\downarrow.
\end{equation}
The minimum reconstruction error is
$\sum_{i>r_g}\lambda_i(\sum_{e\in\mathcal E_g}C^y_e)$.
Thus the shared and temporary expert-specific projections can be solved jointly by an eigendecomposition.

For frequency-weighted fitting, multiply each expert term in the objective by $\tau_e$.
The same derivation then gives
$P_g^\downarrow=\operatorname{eig}_{r_g}(\sum_{e\in\mathcal E_g}\tau_e C^y_e)$, with the same formula for $B_e$.
This solves the weighted objective, which need not minimize the displayed unweighted objective.
Accordingly, $H_g^\downarrow$ in Section~\ref{sec:allocation} uses the unweighted sum to match Obj.~\ref{obj:down_objective}.

\subsection{Down Projections after Intermediate-Width Compression}
\label{app:width_solution}

With $B_e$ fixed, the retained coordinates are $S_e^\top h_e$.
The final expert-specific projection solves
\begin{equation}
    \label{obj:width_fit}
    \min_{A_e^\downarrow}
    \mathbb E_e\|B_eh_e-A_e^\downarrow S_e^\top h_e\|_2^2.
\end{equation}
Its normal equations are
$A_e^\downarrow(S_e^\top C^h_eS_e)=B_eC^h_eS_e$,
giving the Nystr\"om reconstruction in Eq.~(\ref{eq:nystrom}):
$A_e^\downarrow=B_eC^h_eS_e(S_e^\top C^h_eS_e)^\dagger$.
This is a least-squares minimizer for fixed $B_e$ and $S_e$ under the original hidden activations.
It does not jointly optimize neuron selection or account for changes to hidden activations caused by the compressed up/gate projections.
Therefore, the exact down solution before width compression does not imply global optimality of the final compressed expert.

\subsection{Derivation of the Grouping Objective}
\label{app:grouping_derivation}

Fix the candidate shared projections, group ranks, and neuron selections.
Before eliminating the expert-specific projections, assignment minimizes the sum of the up and down reconstruction errors:
\begin{equation}
    \label{obj:grouping_original}
    \begin{aligned}
        \min_{g(\cdot),\{A_e^\uparrow,B_e\}}\quad
        \sum_{e=1}^{E}\mathbb E_e\Big[
         & \|S_e^\top W_e^\uparrow x
        -A_e^\uparrow(P_{g(e)}^\uparrow)^\top x\|_2^2 \\
         & +\|y_e-P_{g(e)}^\downarrow B_eh_e\|_2^2
        \Big].
    \end{aligned}
\end{equation}
The assignment satisfies the same group-size constraints as Obj.~\ref{obj:grouping_objective}.
For the derivation below, consider assigning expert $e$ to a fixed candidate group $g$.
Its optimal projections have dimensions $A_e^\uparrow\in\mathbb R^{k_e\times r_g}$ and $B_e\in\mathbb R^{r_g\times\dff}$.

\paragraph{Squared-error expansion.}
For fixed candidate projections, let $A_e^\uparrow$ and $B_e$ take their least-squares values for each candidate assignment.
Writing $g=g(e)$ within each summand, the assignment-independent terms can be removed:
\begin{equation}
    \label{obj:grouping_expansion}
    \begin{aligned}
            & \operatorname*{arg\,min}_{g(\cdot)}
        \sum_e\mathbb E_e\!\left[
            \|S_e^\top W_e^\uparrow x-A_e^\uparrow(P_g^\uparrow)^\top x\|_2^2
            +\|y_e-P_g^\downarrow B_eh_e\|_2^2
        \right]                                     \\[2pt]
        ={} & \operatorname*{arg\,min}_{g(\cdot)}
        \Bigg\{
        \underbrace{\sum_e\mathbb E_e\!\left[
                \|S_e^\top W_e^\uparrow x\|_2^2+\|y_e\|_2^2
        \right]}_{\text{independent of assignment}} \\[-2pt]
            & \qquad+
        \underbrace{\sum_e\mathbb E_e\!\left[
                \begin{aligned}
                     & \|A_e^\uparrow(P_g^\uparrow)^\top x\|_2^2
                    +\|P_g^\downarrow B_eh_e\|_2^2                                                \\[-2pt]
                     & -2\langle S_e^\top W_e^\uparrow x,A_e^\uparrow(P_g^\uparrow)^\top x\rangle
                    -2\langle y_e,P_g^\downarrow B_eh_e\rangle
                \end{aligned}
                \right]}_{\text{dependent on assignment}}
        \Bigg\}                                     \\[2pt]
        ={} & \operatorname*{arg\,max}_{g(\cdot)}
        \sum_e\underbrace{\mathbb E_e\!\left[
                \begin{aligned}
                     & 2\langle S_e^\top W_e^\uparrow x,A_e^\uparrow(P_g^\uparrow)^\top x\rangle
                    +2\langle y_e,P_g^\downarrow B_eh_e\rangle                                   \\[-2pt]
                     & -\|A_e^\uparrow(P_g^\uparrow)^\top x\|_2^2
                    -\|P_g^\downarrow B_eh_e\|_2^2
                \end{aligned}
                \right]}_{J_e^g}.
    \end{aligned}
\end{equation}
The remaining per-expert score is $J_e^g$; the following steps evaluate it using the calibration second moments.

\paragraph{Up reconstruction error.}
Expanding the expectation gives
\begin{equation}
    \label{eq:grouping_up_expansion}
    \begin{aligned}
         & \mathbb E_e\|S_e^\top W_e^\uparrow x-A_e^\uparrow(P_g^\uparrow)^\top x\|_2^2  \\
         & \quad=\tr\!\left[S_e^\top W_e^\uparrow C^x_e(W_e^\uparrow)^\top S_e\right]
        -2\tr\!\left[A_e^\uparrow(P_g^\uparrow)^\top C^x_e(W_e^\uparrow)^\top S_e\right] \\
         & \qquad+\tr\!\left[A_e^\uparrow(P_g^\uparrow)^\top
            C^x_eP_g^\uparrow(A_e^\uparrow)^\top\right].
    \end{aligned}
\end{equation}
At the least-squares solution in Eq.~(\ref{eq:Aup}), the normal equations make the final trace equal to the cross trace.
Substitution therefore gives
\begin{equation}
    \label{obj:grouping_up_minimum}
    \begin{aligned}
         & \min_{A_e^\uparrow}
        \mathbb E_e\|S_e^\top W_e^\uparrow x-A_e^\uparrow(P_g^\uparrow)^\top x\|_2^2  \\
         & \quad=\tr\!\left[S_e^\top W_e^\uparrow C^x_e(W_e^\uparrow)^\top S_e\right] \\
         & \qquad-\tr\!\Big[
            \left((P_g^\uparrow)^\top C^x_eP_g^\uparrow\right)^{-1}
            (P_g^\uparrow)^\top C^x_e(W_e^\uparrow)^\top
            S_eS_e^\top W_e^\uparrow C^x_eP_g^\uparrow
            \Big].
    \end{aligned}
\end{equation}
The first term depends only on the original expert and selected neurons; the second depends on the candidate shared projection.

\paragraph{Down reconstruction error.}
The orthogonal decomposition in Eq.~(\ref{eq:down_error_split}) and the solution $B_e=(P_g^\downarrow)^\top W_e^\downarrow$ give
\begin{equation}
    \label{obj:grouping_down_minimum}
    \min_{B_e}\mathbb E_e\|y_e-P_g^\downarrow B_eh_e\|_2^2
    =\tr(C^y_e)-\tr\!\left[(P_g^\downarrow)^\top C^y_eP_g^\downarrow\right].
\end{equation}
Again, the first term is independent of the candidate group.

\paragraph{Assignment after elimination.}
Summing the two minimized errors over experts gives $\xi-\sum_e J_e^{g(e)}$, where
\begin{equation}
    \label{eq:grouping_decomposition}
    \begin{aligned}
        \xi={}   & \sum_e\left\{
        \tr\!\left[S_e^\top W_e^\uparrow C^x_e(W_e^\uparrow)^\top S_e\right]
        +\tr(C^y_e)\right\},                                                        \\
        J_e^g={} & \tr\!\Big[
            \left((P_g^\uparrow)^\top C^x_eP_g^\uparrow\right)^{-1}
        (P_g^\uparrow)^\top C^x_e(W_e^\uparrow)^\top                                \\
                 & \hspace{2.8em}{}\cdot S_eS_e^\top W_e^\uparrow C^x_eP_g^\uparrow
            \Big]
        +\tr\!\left[(P_g^\downarrow)^\top C^y_eP_g^\downarrow\right].
    \end{aligned}
\end{equation}
The constant $\xi$ is independent of the assignment, so minimizing the summed error has the same assignments as maximizing $\sum_e J_e^{g(e)}$.
Expanding the squared norms directly gives the four-term expression in Obj.~\ref{obj:grouping_expansion}; evaluating it at the optimal expert-specific projections yields Obj.~\ref{obj:grouping_objective}.
Thus, the minimizers of Obj.~\ref{obj:grouping_original} over assignments coincide with those in Obj.~\ref{obj:grouping_objective} after the expert-specific projections are fitted for each candidate group.
For gated experts, the same calculation adds the gate reconstruction error of a zero prediction to $\xi$ and its error reduction to $J_e^g$.
The equivalence assumes fixed candidate projections and selections and does not guarantee improvement after surrogate refitting.
For the balanced grouping score in Section~\ref{sec:grouping}, multiplying the down reconstruction error by the fixed $\beta_{\mathrm{scale}}$ multiplies both its assignment-independent term and its error reduction by the same factor; the elimination and assignment equivalence remain valid for this weighted objective.

\subsection{Discrete Group-Rank Allocation}
\label{app:rank_solver}

For fixed groups and neuron selections, Obj.~\ref{obj:rank_alloc} has separable scores $V_g$ and per-rank costs $c_g$.
Searching a multiplier for a shared resource constraint is a standard approach in continuous resource allocation \citep{bib:Patriksson2008}; water-filling provides a related continuous example \citep[Example~5.2]{bib:Boyd2004}.
These references motivate the multiplier search, rather than establish optimality of the discrete procedure below.

For a multiplier $\lambda\geq0$, each group independently chooses an integer rank maximizing $V_g(t)-\lambda\cdot c_g\cdot t$ over $r_{\min}\leq t\leq r_{\max}$.
Because the eigenvalues defining $V_g$ are non-increasing, successive rank increments have non-increasing gains.
Choosing the smaller rank in a tie makes total parameter cost non-increasing in $\lambda$.
Bisection increases $\lambda$ when the total cost exceeds $r\cdot\sum_g c_g$ and decreases it when the cost is below that budget, retaining a feasible allocation.
This assumes that the minimum allowed ranks fit within the budget.

Starting from a feasible allocation, residual budget can be assigned by repeatedly selecting the affordable next rank increment with the largest gain per parameter,
$[V_g(r_g+1)-V_g(r_g)]/c_g$, subject to the rank upper bounds.
The procedure stops when no positive-gain increment fits.
Different $c_g$ and integer ranks can leave budget slack, and this multiplier search with greedy completion does not generally guarantee a global optimum of Obj.~\ref{obj:rank_alloc}.

\subsection{Layer Allocation and Budget Correction}
\label{app:layer_allocation}

This section details the across-layer allocation in Section~\ref{sec:allocation}: selecting the overall rank--width split, distributing compression by layer importance, and correcting the total parameter budget.
\paragraph{Layer budgets and shared overhead.}
For layers with the same $n$, $E$, $d$, $\dff$, and $G$, an uncompressed layer holds $n\cdot E\cdot d\cdot\dff$ routed parameters. At reference rank $r_\ell$, the shared projections hold $n\cdot G\cdot d\cdot r_\ell$ parameters and the expert-specific projections hold $n\cdot r_\ell\cdot K_\ell$. The reference retained fraction is
therefore
$\mathcal P_\ell=\alpha_\ell\cdot(\beta_\ell+c_{\text{shared}})$, where $\alpha_\ell=r_\ell/d$, $\beta_\ell=K_\ell/(E\cdot\dff)$, and $c_{\text{shared}}=G\cdot d/(E\cdot\dff)$,
with rank retention $\alpha_\ell$, width retention $\beta_\ell$, and overhead
$c_{\text{shared}}$ of the shared bases. We allocate in terms of $(\alpha_\ell,\beta_\ell)$ and
recover integer budgets by rounding and clipping $r_\ell=\lfloor\alpha_\ell\cdot d\rfloor$
and $K_\ell=\lfloor\beta_\ell\cdot E\cdot\dff\rfloor$ to the prescribed rank and width bounds.
Here $r_\ell$ is the reference rank of Obj.~\ref{obj:rank_alloc} and $K_\ell=\sum_e k_e$.
Since $\sum_g c_g=G\cdot d+K_\ell$, Obj.~\ref{obj:rank_alloc} bounds the actual layer parameter count by this reference budget; discrete allocation can leave unused slack.
\paragraph{Global split.}
Let $\rho$ be the target reduction in total model parameters, and let $N$ and $N_{\mathrm{routed}}$ denote the original total and routed-expert parameter counts, respectively.
Since parameters outside the routed experts remain unchanged, the routed-expert budget is $N_{\mathrm{routed}}-\rho\cdot N$, giving the retained fraction $R=1-\rho\cdot N/N_{\mathrm{routed}}$.
We divide this routed-expert retention between width and rank with an exponent $\nu\in[0,1)$:
$\bar\beta=R^\nu$ and $\bar\alpha=R/(\bar\beta+c_{\text{shared}})$,
so that $\bar\alpha(\bar\beta+c_{\text{shared}})=R$ under uniform allocation, with $\nu=0$
recovering pure factorization ($\bar\beta=1$). Pure width pruning is a separate configuration in
which the shared bases are dropped altogether ($\alpha_\ell\equiv1$, $c_{\text{shared}}=0$,
$\bar\beta=R$); it is not the $\nu\to1$ limit of this formula.

For $L$ layers with positive importance scores $I=(I_1,\ldots,I_L)$, we distribute width and rank removal using $p^{(a)}=\operatorname{softmax}(-\log_{10}I/\epsilon_a)$ for $a\in\{w,r\}$, with temperature $\epsilon_a>0$.
Clipping the per-layer removals $L\cdot(1-\bar\beta)\cdot p_\ell^{(w)}$ and $L\cdot(1-\bar\alpha)\cdot p_\ell^{(r)}$ and redistributing excess removal gives preliminary retentions $\hat\beta_\ell$ and $\hat\alpha_\ell$.
We keep $\beta_\ell=\hat\beta_\ell$ and rescale ranks as $\alpha_\ell=\kappa\cdot\hat\alpha_\ell$, where $\kappa=R/[\frac1L\sum_m\hat\alpha_m\cdot(\hat\beta_m+c_{\text{shared}})]$, to obtain $\frac1L\sum_\ell\mathcal P_\ell=R$ in the continuous relaxation, provided the preliminary allocations and rescaled ranks are feasible.
Rounding and discrete allocation can change the achieved compression.

\section{Additional Experimental Settings}
\label{app:experimental_settings}

We give the numerical and allocation details for the experiments in Section~\ref{sec:settings}.
Neuron selection and projection fitting follow Section~\ref{sec:factorization}, expert grouping follows Section~\ref{sec:grouping}, and capacity allocation follows Section~\ref{sec:allocation}.
\paragraph{Neuron selection.}
We retain at least $5\%$ of the original intermediate width per expert, with the layer width floor set to $E\cdot\lfloor0.05\cdot\dff\rfloor$.
The ridge parameter is selected from $\eta\in\{0.01,0.1\}$ according to model type and calibration size.

Unless otherwise specified, Mixtral-8x7B uses $\eta=0.01$, while DeepSeek-MoE-16B-Base and Qwen2-57B-A14B use $\eta=0.1$, with 512 Alpaca calibration samples of 2048 tokens each.

\paragraph{Experts with few routed calibration inputs.}
For experts with fewer than $2\cdot\dff$ hidden-activation or hidden-state samples, we feed randomly sampled hidden-states and activation batches until at least $2\cdot\dff$ samples are collected.
Experts with fewer than $2\cdot\dff$ routed input tokens are excluded from the shared update and use the fallback
$A_e^\uparrow=S_e^\top W_e^\uparrow C_gP_g^\uparrow
    ((P_g^\uparrow)^\top C_gP_g^\uparrow)^{-1}$.
These experts receive widths close to the minimum because their aggregate neuron leverage scores are smaller.
Qwen2-57B-A14B contains experts with zero routed calibration tokens; the other evaluated models have sufficiently routed experts.

\paragraph{Input-statistics regularization.}
For numerical computation, we use
$C_g=\sum_{e\in\mathcal E_g}\tau_e
    (C_e^x+\eta_x\cdot\tr(C_e^x)\cdot I)$ with $\eta_x=1.0$.
This regularized matrix replaces $C_g$ in the up surrogate construction.
\paragraph{Grouping.}
Balanced medoid clustering uses $k$-means++ seeding with $8$ restarts and at most $25$ Lloyd iterations.
The pairwise gain sweep is computed in FP32.

\paragraph{Rank and layer allocation.}
Group ranks satisfy
$r_{\min}=\max(\lfloor0.35\cdot r_\ell\rfloor,8)$ and
$r_{\max}=\min(\lfloor2.5\cdot r_\ell\rfloor,d)$.
Layer reference ranks are recovered as
$r_\ell=\operatorname{clip}(\lfloor\alpha_\ell\cdot d\rfloor,8,d)$,
and total widths as
$K_\ell=\operatorname{clip}(\lfloor\beta_\ell\cdot E\cdot\dff\rfloor,
    E\cdot\lfloor0.05\cdot\dff\rfloor,E\cdot\dff)$.
For either allocation axis, layer removal fractions are clipped to $[0.02,\omega]$,
where $\omega=0.85$, increasing to $0.95$ when the corresponding aggregate removal exceeds $0.40$.
Clipped mass is redistributed over layers that have not reached their bounds.
We report achieved compression after integer rounding and discrete rank allocation.

\begin{table}[t]
    \centering
    \caption{Default compression hyperparameters for the main comparisons. Ablation settings are specified separately.}
    \label{tab:compression_hyperparameters}
    \begin{tabular}{lcccc}
        \toprule
        Model    & Compression Rates & $\eta$ & $\nu$ & $G$ \\
        \midrule
        DeepSeek & 20\%              & 0.1    & 0.75  & 1   \\
                 & 40\%              & 0.1    & 0.65  & 2   \\
                 & 60\%              & 0.1    & 0.55  & 8   \\
        \midrule
        Mixtral  & 20\%              & 0.01   & 0.85  & 1   \\
                 & 40\%              & 0.01   & 0.85  & 1   \\
                 & 60\%              & 0.01   & 0.85  & 1   \\
        \midrule
        Qwen     & 20\%              & 0.1    & 0.75  & 1   \\
                 & 40\%              & 0.1    & 0.65  & 2   \\
                 & 60\%              & 0.1    & 0.55  & 8   \\
        \bottomrule
    \end{tabular}

    \smallskip
    \begin{minipage}{\linewidth}
        \footnotesize
        For DeepSeek and Qwen with 256 calibration samples,
        $\eta=0.01$ was more effective; the main comparisons with 512 calibration samples
        use $\eta=0.1$ for these models.
    \end{minipage}
\end{table}

\section{Additional Experiments}
\label{app:additional_experiments}

\begin{figure}[h]
    \centering
    \begin{subfigure}[b]{0.49\linewidth}
        \centering
        \includegraphics[width=\linewidth]{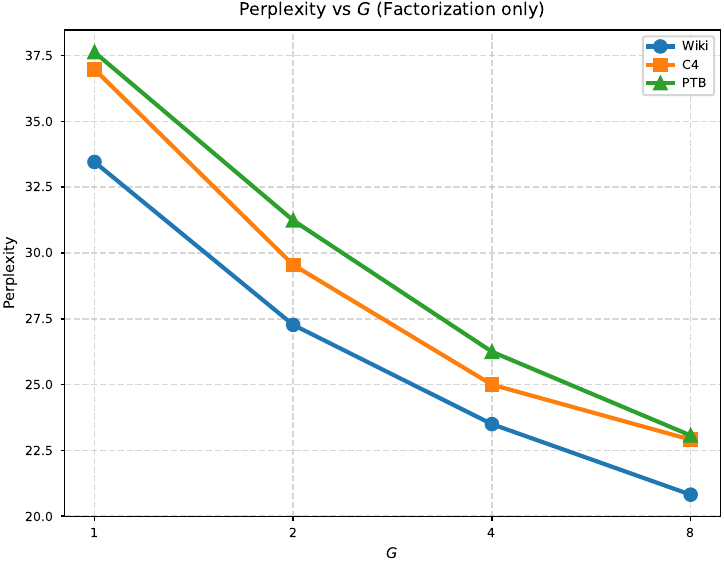}
        \caption{Effect of $G \in \{1, 2, 4, 8\}$.}
        \label{fig:G_ppl}
    \end{subfigure}
    \hfill
    \begin{subfigure}[b]{0.49\linewidth}
        \centering
        \includegraphics[width=\linewidth]{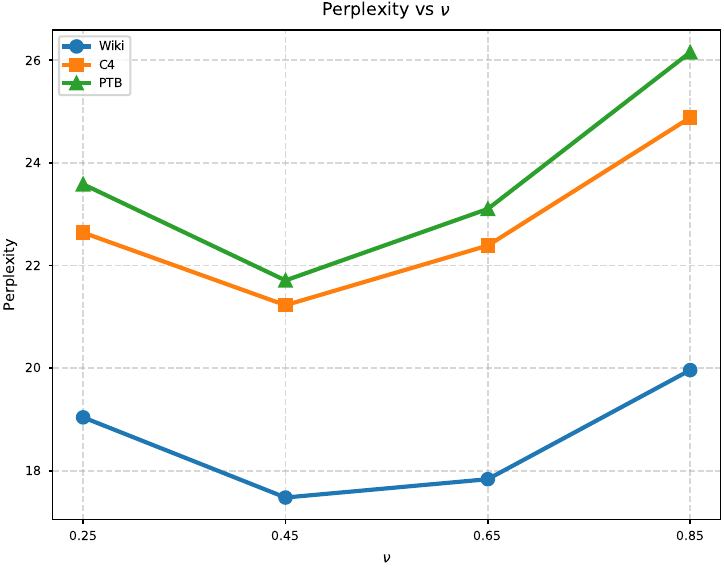}
        \caption{Effect of $\nu$ at four broadly spaced values.}
        \label{fig:nu_ppl}
    \end{subfigure}
    \caption{Perplexity of GS-MoE on DeepSeek-16B-Base at 55\% compression. \textbf{(a)} Varying the number of groups $G$. \textbf{(b)} Varying $\nu$.}
    \label{fig:G_nu_ppl}
\end{figure}

\paragraph{Effect of $G$ and $\nu$.}
Figure ~\ref{fig:G_nu_ppl} (left) varies the number of groups with factorization leaving intermediate width full-rank. Perplexity decreases monotonically with $G$ on all three datasets, with WikiText falling from 33.5 at $G=1$ to 20.75 at $G=8$. Returns diminish with each doubling (6.1, 3.8, and 2.7 points on WikiText). This is consistent with outlier-experts being more capable of representing features under smaller group sizes. Figure ~\ref{fig:G_nu_ppl} (right) varies width retention by the hyperparameter $\nu$ detailed in ~\ref{app:layer_allocation}. Minimum perplexity is found at $\nu=0.45$, with additional increments increasing perplexity. Smaller $\nu$ values restrict expert representation, degrading per-expert resolution, while large values of $\nu$ allocate more subspace rank without the proper compensatory compression rate to maximize accuracy.

\input{Sections/Tables/NonUniformExpWidth}

\subsection{Non-Uniform Layer and Intermediate Width Allocation}
We compare non-uniform and uniform expert width and layer allocation in Table ~\ref{tab:allocation}. Table ~\ref{tab:allocation} demonstrates improvement under non-uniform per-expert intermediate width allocation for Qwen models, improving Wikitext perplexity by 0.23 at 60\% compression. Similarly, non-uniform layer-wise allocation reduces Wikitext perplexity by 0.26 at 60\% compression.

\input{Sections/Tables/GSMoESubsets}

\subsection{Evaluations of GS-MoE's Subsets}

Table \ref{tab:factorization_nystrom} provides further insight on GS-MoE's performance against fine-grained and coarse-grained architectures.
Table \ref{tab:factorization_nystrom} lists three competing methods: pure factorization, pure width reduction (Nystr\"om), and the full pipeline (GS-MoE).
For Qwen models, pure factorization performs in-line with width reduction while at higher compression ratios the two methods alone drop significantly in performance.
By incorporating the full-pipeline of GS-MoE, GS-MoE is able to recover the performance loss that occurs at 60\% compression of either method.
The Deepseek architecture tells a similar story, demonstrating that combining GS-MoE's grouped-subspace factorization with pruning pushes the compression limit higher than what other methods offer.
Next, GS-MoE's expert width reduction is effectively able to compress Mixtral-8x7B up to 40\% compression, performing better than GS-MoE's factorization approach at the same ratios.
However, GS-MoE's full-pipeline struggles to exceed the reasoning performance of width-reduction.
We attribute this to the larger intermediate dimension of Mixtral-8x7B and reduced expert count, making subspace rank reduction more damaging and limiting the efficacy of GS-MoE's expert subspace assignment.

%% file: Sections/Tables/NonUniformExpWidth.tex
\begin{table}[!t]
\centering
\caption{Perplexity ($\downarrow$) of GS-MoE on Qwen2-57B-A14B under uniform and non-uniform allocation.
\textbf{(a)} Uniform vs.\ leverage-based expert widths; $\nu$ controls the split between width and rank compression.
\textbf{(b)} Uniform vs.\ non-uniform layer-wise allocation, calibrated on Alpaca.
Lower perplexity for each dataset and setting is in \textbf{bold}.}
\label{tab:allocation}
\small
\setlength{\tabcolsep}{4pt}
\begin{tabular}{cc@{\hspace{14pt}}ccc@{\hspace{14pt}}ccc}
\toprule
\multicolumn{8}{l}{\textit{(a) Per-expert width allocation}} \\
\midrule
& & \multicolumn{3}{c}{Uniform} & \multicolumn{3}{c}{Non-uniform} \\
\cmidrule(lr){3-5}\cmidrule(lr){6-8}
Ratio & $\nu$ & Wiki & C4 & PTB & Wiki & C4 & PTB \\
\midrule
20\% & 0.75 & 6.17 & 9.08  & 9.33  & \textbf{6.07} & \textbf{8.98}  & \textbf{9.26}  \\
40\% & 0.65 & 6.83 & 10.16 & 10.05 & \textbf{6.59} & \textbf{10.02} & \textbf{9.81}  \\
60\% & 0.55 & 8.10 & 12.15 & 11.65 & \textbf{7.87} & \textbf{11.93} & \textbf{11.29} \\
\midrule
\multicolumn{8}{l}{\textit{(b) Layer-wise allocation}} \\
\midrule
& & \multicolumn{3}{c}{Uniform} & \multicolumn{3}{c}{Non-uniform} \\
\cmidrule(lr){3-5}\cmidrule(lr){6-8}
Ratio & $\nu$ & Wiki & C4 & PTB & Wiki & C4 & PTB \\
\midrule
60\% & 0.55 & 8.12 & 12.34 & 11.56 & \textbf{7.87} & \textbf{11.93} & \textbf{11.29} \\
\bottomrule
\end{tabular}
\end{table}

%% file: Sections/Tables/GSMoESubsets.tex
\begin{table*}[!t]
\caption{Subsets of GS-MoE on three different models. Factorization is GS-MoE's group shared subspace method with full intermediate width. Nystr\"om is GS-MoE with only expert intermediate width reduction.}
\label{tab:factorization_nystrom}
\vspace{3pt}
\centering
\scriptsize
\setlength{\tabcolsep}{6pt}
\renewcommand{\arraystretch}{1.1}
\setlength{\arrayrulewidth}{0.4pt}
\resizebox{\ifdim\width>\linewidth\linewidth\else\width\fi}{!}{%
\begin{tabular}{c|l|rr|rrrrrrrr}
\noalign{\global\arrayrulewidth=0.8pt}\hline\noalign{\global\arrayrulewidth=0.4pt}
Ratio & Method & Wiki.$\downarrow$ & PTB$\downarrow$ & OpenB. & ARC-e & WinoG. & HellaS. & ARC-c & PIQA & MathQA & Avg.$\uparrow$ \\
\hline
\multicolumn{12}{c}{\textbf{DeepSeekMoE-16B-Base} (64 experts, top-6 activated)} \\
\hline
0\% & Original & 6.38 & 9.47 & 0.32 & 0.75 & 0.70 & 0.58 & 0.44 & 0.79 & 0.31 & 0.56 \\
\hline
\multirow{3}{*}{20\%}
 & Factorization & 8.11 & 11.47 & 0.30 & 0.74 & \textbf{0.70} & 0.54 & 0.43 & 0.77 & 0.28 & 0.538 \\
 & Nystr\"om & \textbf{7.30} & \textbf{10.53} & \textbf{0.33} & \textbf{0.76} & \textbf{0.70} & \textbf{0.57} & \textbf{0.44} & \textbf{0.79} & \textbf{0.32} & \textbf{0.558} \\
 & \textbf{GS-MoE (Ours)} & 7.35 & 10.64 & \textbf{0.33} & 0.75 & \textbf{0.70} & \textbf{0.57} & 0.43 & \textbf{0.79} & 0.30 & 0.552 \\
\hline
\multirow{3}{*}{40\%}
 & Factorization & 12.97 & 16.13 & 0.27 & 0.71 & 0.68 & 0.48 & 0.39 & 0.74 & 0.26 & 0.506 \\
 & Nystr\"om & \textbf{10.09} & 14.30 & 0.30 & \textbf{0.75} & 0.69 & \textbf{0.53} & 0.42 & \textbf{0.78} & \textbf{0.31} & 0.539 \\
 & \textbf{GS-MoE (Ours)} & 10.10 & \textbf{13.51} & \textbf{0.31} & \textbf{0.75} & \textbf{0.70} & \textbf{0.53} & \textbf{0.43} & \textbf{0.78} & 0.29 & \textbf{0.543} \\ %
\hline
\multirow{3}{*}{60\%}
 & Factorization & 25.02 & 29.37 & 0.24 & 0.66 & 0.64 & 0.39 & 0.35 & 0.70 & 0.25 & 0.462 \\
 & Nystr\"om & 18.62 & 27.12 & 0.27 & 0.71 & 0.62 & 0.41 & 0.37 & 0.73 & \textbf{0.29} & 0.487 \\
 & \textbf{GS-MoE (Ours)} & \textbf{17.55} & \textbf{22.21} & \textbf{0.28} & \textbf{0.72} & \textbf{0.66} & \textbf{0.45} & \textbf{0.39} & \textbf{0.74} & 0.28 & \textbf{0.503} \\
\hline
\multicolumn{12}{c}{\textbf{Qwen2-57B-A14B} (64 experts, top-8 activated)} \\
\hline
0\% & Original & 5.12 & 9.18 & 0.33 & 0.75 & 0.74 & 0.63 & 0.46 & 0.81 & 0.39 & 0.59 \\
\hline
\multirow{3}{*}{40\%}
 & Factorization & 7.19 & 10.46 & 0.31 & \textbf{0.78} & \textbf{0.76} & 0.59 & \textbf{0.49} & 0.80 & \textbf{0.40} & 0.591 \\
 & Nystr\"om & 6.61 & 9.83 & 0.33 & \textbf{0.78} & 0.75 & \textbf{0.62} & 0.48 & \textbf{0.81} & 0.39 & \textbf{0.593} \\
 & \textbf{GS-MoE (Ours)} & \textbf{6.59} & \textbf{9.81} & \textbf{0.34} & 0.77 & 0.75 & \textbf{0.62} & 0.47 & \textbf{0.81} & 0.39 & 0.592 \\
\hline
\multirow{3}{*}{60\%}
 & Factorization & 9.61 & 14.47 & 0.29 & 0.77 & 0.75 & 0.54 & 0.47 & 0.79 & 0.36 & 0.567 \\
 & Nystr\"om & 7.89 & 11.47 & \textbf{0.32} & 0.78 & 0.73 & \textbf{0.60} & 0.49 & \textbf{0.80} & \textbf{0.38} & 0.586 \\
 & \textbf{GS-MoE (Ours)} & \textbf{7.87} & \textbf{11.29} & \textbf{0.32} & \textbf{0.79} & \textbf{0.78} & 0.59 & \textbf{0.50} & \textbf{0.80} & \textbf{0.38} & \textbf{0.594} \\ %
\hline
\multicolumn{12}{c}{\textbf{Mixtral-8$\times$7B-v0.1} (8 experts, top-2 activated)} \\
\hline
0\% & Original & 3.98 & 12.99 & 0.36 & 0.84 & 0.76 & 0.65 & 0.57 & 0.82 & 0.43 & 0.63 \\
\hline
\multirow{3}{*}{20\%}
 & Factorization & 6.25 & \textbf{15.64} & 0.34 & 0.81 & \textbf{0.76} & 0.59 & 0.51 & 0.81 & 0.38 & 0.599 \\
 & Nystr\"om & \textbf{5.81} & 16.05 & \textbf{0.36} & \textbf{0.82} & 0.75 & \textbf{0.62} & \textbf{0.53} & \textbf{0.82} & \textbf{0.41} & \textbf{0.618} \\
 & \textbf{GS-MoE (Ours)} & 5.88 & 15.69 & \textbf{0.36} & \textbf{0.82} & \textbf{0.76} & \textbf{0.62} & \textbf{0.53} & \textbf{0.82} & \textbf{0.41} & 0.616 \\
\hline
\multirow{3}{*}{40\%}
 & Factorization & 20.00 & 35.08 & 0.28 & 0.73 & 0.70 & 0.45 & 0.41 & 0.74 & 0.30 & 0.514 \\
 & Nystr\"om & 9.56 & 22.35 & \textbf{0.31} & \textbf{0.79} & \textbf{0.71} & \textbf{0.55} & 0.48 & \textbf{0.79} & \textbf{0.39} & \textbf{0.574} \\
 & \textbf{GS-MoE (Ours)} & \textbf{9.29} & \textbf{22.17} & \textbf{0.31} & \textbf{0.79} & 0.70 & \textbf{0.55} & \textbf{0.49} & \textbf{0.79} & 0.38 & 0.572 \\
\hline
\multirow{3}{*}{60\%}
 & Factorization & 62.10 & 103.50 & 0.19 & 0.51 & 0.57 & 0.34 & 0.27 & 0.64 & 0.24 & 0.394 \\
 & Nystr\"om & 16.37 & 37.17 & \textbf{0.26} & 0.70 & 0.63 & 0.43 & \textbf{0.41} & \textbf{0.73} & \textbf{0.35} & \textbf{0.502} \\
 & \textbf{GS-MoE (Ours)} & \textbf{15.94} & \textbf{35.96} & \textbf{0.26} & \textbf{0.72} & \textbf{0.64} & \textbf{0.44} & \textbf{0.41} & \textbf{0.73} & 0.32 & \textbf{0.502} \\
\noalign{\global\arrayrulewidth=0.8pt}\hline\noalign{\global\arrayrulewidth=0.4pt}
\end{tabular}%
}
\end{table*}

%% file: References.bib
@inproceedings{bib:NystromMatrix,
  title     = {Using the {Nystr{\"o}m} Method to Speed Up Kernel Machines},
  author    = {Williams, Christopher K. I. and Seeger, Matthias},
  booktitle = {Advances in Neural Information Processing Systems},
  volume    = {13},
  year      = {2000},
  url       = {https://proceedings.neurips.cc/paper/2000/hash/19de10adbaa1b2ee13f77f679fa1483a-Abstract.html}
}

@inproceedings{bib:MoBE,
  title     = {{MoBE}: Mixture-of-Basis-Experts for Compressing {MoE}-based {LLM}s},
  author    = {Chen, Xiaodong and Ha, Mingming and Lan, Zhenzhong and Zhang, Jing and Li, Jianguo},
  booktitle = {The Fourteenth International Conference on Learning Representations},
  year      = {2026},
  url       = {https://openreview.net/forum?id=8RV6H50OSf}
}

@inproceedings{bib:MoE-SVD,
  title     = {{M}o{E}-{SVD}: Structured Mixture-of-Experts {LLM}s Compression via Singular Value Decomposition},
  author    = {Li, Wei and Li, Lujun and Gu, Hao and Huang, You-Liang and Lee, Mark G. and Sun, Shengjie and Xue, Wei and Guo, Yike},
  booktitle = {Proceedings of the 42nd International Conference on Machine Learning},
  pages     = {35209--35230},
  year      = {2025},
  editor    = {Singh, Aarti and Fazel, Maryam and Hsu, Daniel and Lacoste-Julien, Simon and Berkenkamp, Felix and Maharaj, Tegan and Wagstaff, Kiri and Zhu, Jerry},
  volume    = {267},
  series    = {Proceedings of Machine Learning Research},
  month     = {13--19 Jul},
  publisher = {PMLR},
  url       = {https://proceedings.mlr.press/v267/li25az.html}
}

@inproceedings{bib:D2-MoE,
  title     = {Delta Decompression  for MoE-based {LLM}s Compression},
  author    = {Hao Gu and Wei Li and Lujun Li and Qiyuan Zhu and Mark G. Lee and Shengjie Sun and Wei Xue and Yike Guo},
  booktitle = {Forty-second International Conference on Machine Learning},
  year      = {2025},
  url       = {https://openreview.net/forum?id=ziezViPoN1}
}

@inproceedings{bib:NAEE,
  title     = {Not All Experts are Equal: Efficient Expert Pruning and Skipping for Mixture-of-Experts Large Language Models},
  author    = {Lu, Xudong  and
               Liu, Qi  and
               Xu, Yuhui  and
               Zhou, Aojun  and
               Huang, Siyuan  and
               Zhang, Bo  and
               Yan, Junchi  and
               Li, Hongsheng},
  editor    = {Ku, Lun-Wei  and
               Martins, Andre  and
               Srikumar, Vivek},
  booktitle = {Proceedings of the 62nd Annual Meeting of the Association for Computational Linguistics (Volume 1: Long Papers)},
  month     = aug,
  year      = {2024},
  address   = {Bangkok, Thailand},
  publisher = {Association for Computational Linguistics},
  url       = {https://aclanthology.org/2024.acl-long.334/},
  doi       = {10.18653/v1/2024.acl-long.334},
  pages     = {6159--6172}
}

@misc{bib:DeepSpeed,
  title         = {DeepSpeed-MoE: Advancing Mixture-of-Experts Inference and Training to Power Next-Generation AI Scale},
  author        = {Samyam Rajbhandari and Conglong Li and Zhewei Yao and Minjia Zhang and Reza Yazdani Aminabadi and Ammar Ahmad Awan and Jeff Rasley and Yuxiong He},
  year          = {2022},
  eprint        = {2201.05596},
  archiveprefix = {arXiv},
  primaryclass  = {cs.LG},
  url           = {https://arxiv.org/abs/2201.05596}
}

@inproceedings{bib:TENP,
  title     = {{TENP}: Trapezoidal Expert Neuron Pruning For Mixture-of-Experts},
  author    = {He, Jiangyang  and
               Zhu, Shaolin  and
               Xiong, Deyi},
  editor    = {Liakata, Maria  and
               Moreira, Viviane P.  and
               Zhang, Jiajun  and
               Jurgens, David},
  booktitle = {Findings of the {A}ssociation for {C}omputational {L}inguistics: {ACL} 2026},
  month     = jul,
  year      = {2026},
  address   = {San Diego, California, United States},
  publisher = {Association for Computational Linguistics},
  url       = {https://aclanthology.org/2026.findings-acl.1049/},
  doi       = {10.18653/v1/2026.findings-acl.1049},
  pages     = {20911--20925},
  isbn      = {979-8-89176-395-1}
}

@misc{bib:LessIsMoE,
  title         = {Less is MoE: Trimming Experts in Domain-Specialist Language Models},
  author        = {Haoze He and Xinkai Zou and Xuan Jiang and Xingyuan Ding and Ao Qu and Juncheng Billy Li and Heather Miller},
  year          = {2026},
  eprint        = {2606.05538},
  archiveprefix = {arXiv},
  primaryclass  = {cs.LG},
  url           = {https://arxiv.org/abs/2606.05538}
}

@inproceedings{bib:MoDeGPT,
  title     = {MoDe{GPT}: Modular Decomposition for Large Language Model Compression},
  author    = {Chi-Heng Lin and Shangqian Gao and James Seale Smith and Abhishek Patel and Shikhar Tuli and Yilin Shen and Hongxia Jin and Yen-Chang Hsu},
  booktitle = {The Thirteenth International Conference on Learning Representations},
  year      = {2025},
  url       = {https://openreview.net/forum?id=8EfxjTCg2k}
}

@inproceedings{bib:HC-SMoE,
  title     = {Retraining-free Merging of Sparse MoE via Hierarchical Clustering},
  author    = {I-Chun Chen and Hsu-Shen Liu and Wei-Fang Sun and Chen-Hao Chao and Yen-Chang Hsu and Chun-Yi Lee},
  booktitle = {Forty-second International Conference on Machine Learning},
  year      = {2025},
  url       = {https://openreview.net/forum?id=hslOzRxzXL}
}

@inproceedings{bib:TD-MoE,
  title     = {{TD}-MoE: Tensor Decomposition for MoE Models},
  author    = {Xu, Yuebin and Wang, Yanhong and Peng, Xuemei and Zang, Hui and Chen, Minghao and Xia, Pengfei and Wen, Zeyi},
  booktitle = {The Fourteenth International Conference on Learning Representations},
  year      = {2026},
  url       = {https://openreview.net/forum?id=D9cnZNZfxX}
}

@misc{bib:DeepseekAI,
  title         = {DeepSeek-V3 Technical Report},
  author        = {DeepSeek-AI and Aixin Liu and Bei Feng and Bing Xue and Bingxuan Wang and Bochao Wu and Chengda Lu and Chenggang Zhao and Chengqi Deng and Chenyu Zhang and Chong Ruan and Damai Dai and Daya Guo and Dejian Yang and Deli Chen and Dongjie Ji and Erhang Li and Fangyun Lin and Fucong Dai and Fuli Luo and Guangbo Hao and Guanting Chen and Guowei Li and H. Zhang and Han Bao and Hanwei Xu and Haocheng Wang and Haowei Zhang and Honghui Ding and Huajian Xin and Huazuo Gao and Hui Li and Hui Qu and J. L. Cai and Jian Liang and Jianzhong Guo and Jiaqi Ni and Jiashi Li and Jiawei Wang and Jin Chen and Jingchang Chen and Jingyang Yuan and Junjie Qiu and Junlong Li and Junxiao Song and Kai Dong and Kai Hu and Kaige Gao and Kang Guan and Kexin Huang and Kuai Yu and Lean Wang and Lecong Zhang and Lei Xu and Leyi Xia and Liang Zhao and Litong Wang and Liyue Zhang and Meng Li and Miaojun Wang and Mingchuan Zhang and Minghua Zhang and Minghui Tang and Mingming Li and Ning Tian and Panpan Huang and Peiyi Wang and Peng Zhang and Qiancheng Wang and Qihao Zhu and Qinyu Chen and Qiushi Du and R. J. Chen and R. L. Jin and Ruiqi Ge and Ruisong Zhang and Ruizhe Pan and Runji Wang and Runxin Xu and Ruoyu Zhang and Ruyi Chen and S. S. Li and Shanghao Lu and Shangyan Zhou and Shanhuang Chen and Shaoqing Wu and Shengfeng Ye and Shengfeng Ye and Shirong Ma and Shiyu Wang and Shuang Zhou and Shuiping Yu and Shunfeng Zhou and Shuting Pan and T. Wang and Tao Yun and Tian Pei and Tianyu Sun and W. L. Xiao and Wangding Zeng and Wanjia Zhao and Wei An and Wen Liu and Wenfeng Liang and Wenjun Gao and Wenqin Yu and Wentao Zhang and X. Q. Li and Xiangyue Jin and Xianzu Wang and Xiao Bi and Xiaodong Liu and Xiaohan Wang and Xiaojin Shen and Xiaokang Chen and Xiaokang Zhang and Xiaosha Chen and Xiaotao Nie and Xiaowen Sun and Xiaoxiang Wang and Xin Cheng and Xin Liu and Xin Xie and Xingchao Liu and Xingkai Yu and Xinnan Song and Xinxia Shan and Xinyi Zhou and Xinyu Yang and Xinyuan Li and Xuecheng Su and Xuheng Lin and Y. K. Li and Y. Q. Wang and Y. X. Wei and Y. X. Zhu and Yang Zhang and Yanhong Xu and Yanhong Xu and Yanping Huang and Yao Li and Yao Zhao and Yaofeng Sun and Yaohui Li and Yaohui Wang and Yi Yu and Yi Zheng and Yichao Zhang and Yifan Shi and Yiliang Xiong and Ying He and Ying Tang and Yishi Piao and Yisong Wang and Yixuan Tan and Yiyang Ma and Yiyuan Liu and Yongqiang Guo and Yu Wu and Yuan Ou and Yuchen Zhu and Yuduan Wang and Yue Gong and Yuheng Zou and Yujia He and Yukun Zha and Yunfan Xiong and Yunxian Ma and Yuting Yan and Yuxiang Luo and Yuxiang You and Yuxuan Liu and Yuyang Zhou and Z. F. Wu and Z. Z. Ren and Zehui Ren and Zhangli Sha and Zhe Fu and Zhean Xu and Zhen Huang and Zhen Zhang and Zhenda Xie and Zhengyan Zhang and Zhewen Hao and Zhibin Gou and Zhicheng Ma and Zhigang Yan and Zhihong Shao and Zhipeng Xu and Zhiyu Wu and Zhongyu Zhang and Zhuoshu Li and Zihui Gu and Zijia Zhu and Zijun Liu and Zilin Li and Ziwei Xie and Ziyang Song and Ziyi Gao and Zizheng Pan},
  year          = {2025},
  eprint        = {2412.19437},
  archiveprefix = {arXiv},
  primaryclass  = {cs.CL},
  url           = {https://arxiv.org/abs/2412.19437}
}

@inproceedings{bib:REAP,
  title     = {{REAP} the Experts: Why Pruning Prevails for One-Shot MoE compression},
  author    = {Mike Lasby and Ivan Lazarevich and Nish Sinnadurai and Sean Lie and Yani Ioannou and Vithursan Thangarasa},
  booktitle = {The Fourteenth International Conference on Learning Representations},
  year      = {2026},
  url       = {https://openreview.net/forum?id=ukGxWd2aDG}
}

@inproceedings{bib:MoE-I2,
  title     = {{M}o{E}-I$^2$: Compressing Mixture of Experts Models through Inter-Expert Pruning and Intra-Expert Low-Rank Decomposition},
  author    = {Yang, Cheng  and
               Sui, Yang  and
               Xiao, Jinqi  and
               Huang, Lingyi  and
               Gong, Yu  and
               Duan, Yuanlin  and
               Jia, Wenqi  and
               Yin, Miao  and
               Cheng, Yu  and
               Yuan, Bo},
  editor    = {Al-Onaizan, Yaser  and
               Bansal, Mohit  and
               Chen, Yun-Nung},
  booktitle = {Findings of the Association for Computational Linguistics: EMNLP 2024},
  month     = nov,
  year      = {2024},
  address   = {Miami, Florida, USA},
  publisher = {Association for Computational Linguistics},
  url       = {https://aclanthology.org/2024.findings-emnlp.612/},
  doi       = {10.18653/v1/2024.findings-emnlp.612},
  pages     = {10456--10466}
}

@misc{bib:Alpaca,
  author       = {Rohan Taori and Ishaan Gulrajani and Tianyi Zhang and Yann Dubois and Xuechen Li and Carlos Guestrin and Percy Liang and Tatsunori B. Hashimoto },
  title        = {Stanford Alpaca: An Instruction-following LLaMA model},
  year         = {2023},
  publisher    = {GitHub},
  journal      = {GitHub repository},
  howpublished = {\url{https://github.com/tatsu-lab/stanford_alpaca}}
}

@software{bib:LM-Eval,
  author    = {Gao, Leo and
               Tow, Jonathan and
               Biderman, Stella and
               Black, Sid and
               DiPofi, Anthony and
               Foster, Charles and
               Golding, Laurence and
               Hsu, Jeffrey and
               McDonell, Kyle and
               Muennighoff, Niklas and
               Phang, Jason and
               Reynolds, Laria and
               Tang, Eric and
               Thite, Anish and
               Wang, Ben and
               Wang, Kevin and
               Zou, Andy},
  title     = {A framework for few-shot language model evaluation},
  month     = sep,
  year      = 2021,
  publisher = {Zenodo},
  version   = {v0.0.1},
  doi       = {10.5281/zenodo.5371628},
  url       = {https://doi.org/10.5281/zenodo.5371628}
}

@misc{bib:Wikitext,
  title         = {Pointer Sentinel Mixture Models},
  author        = {Stephen Merity and Caiming Xiong and James Bradbury and Richard Socher},
  year          = {2016},
  eprint        = {1609.07843},
  archiveprefix = {arXiv},
  primaryclass  = {cs.CL}
}

@article{bib:C4-dataset,
  author  = {Colin Raffel and Noam Shazeer and Adam Roberts and Katherine Lee and Sharan Narang and Michael Matena and Yanqi Zhou and Wei Li and Peter J. Liu},
  title   = {Exploring the Limits of Transfer Learning with a Unified Text-to-Text Transformer},
  journal = {Journal of Machine Learning Research},
  year    = {2020},
  volume  = {21},
  number  = {140},
  pages   = {1--67},
  url     = {http://jmlr.org/papers/v21/20-074.html}
}

@inproceedings{bib:DERN,
  title     = {Dropping Experts, Recombining Neurons: Retraining-Free Pruning for Sparse Mixture-of-Experts {LLM}s},
  author    = {Zhou, Yixiao and Zhao, Ziyu and Cheng, Dongzhou and Wu, Zhiliang and Gui, Jie and Yang, Yi and Wu, Fei and Cheng, Yu and Fan, Hehe},
  booktitle = {Findings of the Association for Computational Linguistics: EMNLP 2025},
  year      = {2025},
  month     = nov,
  pages     = {15169--15186},
  publisher = {Association for Computational Linguistics},
  doi       = {10.18653/v1/2025.findings-emnlp.820},
  url       = {https://aclanthology.org/2025.findings-emnlp.820/}
}

@misc{bib:MoLAE,
  title         = {{MoLAE}: Mixture of Latent Experts for Parameter-Efficient Language Models},
  author        = {Zehua Liu and Han Wu and Ruifeng She and Xiaojin Fu and Xiongwei Han and Tao Zhong and Mingxuan Yuan},
  year          = {2025},
  eprint        = {2503.23100},
  archiveprefix = {arXiv},
  primaryclass  = {cs.LG},
  url           = {https://arxiv.org/abs/2503.23100}
}

@misc{bib:Sub-MoE,
  title         = {{Sub-MoE}: Efficient Mixture-of-Expert {LLM}s Compression via Subspace Expert Merging},
  author        = {Lujun Li and Qiyuan Zhu and Jiacheng Wang and Wei Li and Hao Gu and Sirui Han and Yike Guo},
  year          = {2025},
  eprint        = {2506.23266},
  archiveprefix = {arXiv},
  primaryclass  = {cs.LG},
  url           = {https://arxiv.org/abs/2506.23266}
}

@misc{bib:LorExperts,
  title         = {Shape Mutating Expert Compression: {LorExperts} and {BTExperts}},
  author        = {Inesh Chakrabarti and Sourjya Roy and Bowen Bao and Thiago Crepaldi and Spandan Tiwari and Ashish Sirasao},
  year          = {2026},
  eprint        = {2608.07814},
  archiveprefix = {arXiv},
  primaryclass  = {cs.LG},
  url           = {https://arxiv.org/abs/2608.07814}
}

@article{bib:JonkerVolgenant,
  author   = {Crouse, David F.},
  journal  = {IEEE Transactions on Aerospace and Electronic Systems},
  title    = {On implementing 2D rectangular assignment algorithms},
  year     = {2016},
  volume   = {52},
  number   = {4},
  pages    = {1679-1696},
  doi      = {10.1109/TAES.2016.140952}
}

@article{bib:Patriksson2008,
  author  = {Michael Patriksson},
  title   = {A survey on the continuous nonlinear resource allocation problem},
  journal = {European Journal of Operational Research},
  year    = {2008},
  volume  = {185},
  number  = {1},
  pages   = {1--46},
  doi     = {10.1016/j.ejor.2006.12.006}
}

@article{bib:Lloyd1982,
  author  = {Stuart P. Lloyd},
  title   = {Least Squares Quantization in {PCM}},
  journal = {IEEE Transactions on Information Theory},
  volume  = {28},
  number  = {2},
  pages   = {129--137},
  year    = {1982},
  doi     = {10.1109/TIT.1982.1056489}
}

@inproceedings{bib:Arthur2007,
  author    = {David Arthur and Sergei Vassilvitskii},
  title     = {$k$-means++: The Advantages of Careful Seeding},
  booktitle = {Proceedings of the Eighteenth Annual ACM-SIAM Symposium on Discrete Algorithms},
  pages     = {1027--1035},
  year      = {2007},
  url       = {https://theory.stanford.edu/~sergei/papers/kMeansPP-soda.pdf}
}

@book{bib:Boyd2004,
  author    = {Stephen Boyd and Lieven Vandenberghe},
  title     = {Convex Optimization},
  publisher = {Cambridge University Press},
  year      = {2004},
  url       = {https://web.stanford.edu/~boyd/cvxbook/}
}

@inproceedings{bib:OpenbookQA,
  title     = {Can a Suit of Armor Conduct Electricity? A New Dataset for Open Book Question Answering},
  author    = {Todor Mihaylov and Peter Clark and Tushar Khot and Ashish Sabharwal},
  booktitle = {EMNLP},
  year      = {2018}
}

@inproceedings{bib:WinoGrande,
  title  = {WinoGrande: An Adversarial Winograd Schema Challenge at Scale},
  author = {Keisuke, Sakaguchi and Ronan, Le Bras and Chandra, Bhagavatula and Yejin, Choi},
  year   = {2019}
}

@inproceedings{bib:HellaSwag,
  title     = {HellaSwag: Can a Machine Really Finish Your Sentence?},
  author    = {Zellers, Rowan and Holtzman, Ari and Bisk, Yonatan and Farhadi, Ali and Choi, Yejin},
  booktitle = {Proceedings of the 57th Annual Meeting of the Association for Computational Linguistics},
  year      = {2019}
}

@inproceedings{bib:PIQA,
  author    = {Yonatan Bisk and Rowan Zellers and
               Ronan Le Bras and Jianfeng Gao
               and Yejin Choi},
  title     = {PIQA: Reasoning about Physical Commonsense in
               Natural Language},
  booktitle = {Thirty-Fourth AAAI Conference on
               Artificial Intelligence},
  year      = {2020}
}

@inproceedings{bib:MathQA,
  title     = {{M}ath{QA}: Towards Interpretable Math Word Problem Solving with Operation-Based Formalisms},
  author    = {Amini, Aida  and
               Gabriel, Saadia  and
               Lin, Shanchuan  and
               Koncel-Kedziorski, Rik  and
               Choi, Yejin  and
               Hajishirzi, Hannaneh},
  booktitle = {Proceedings of the 2019 Conference of the North {A}merican Chapter of the Association for Computational Linguistics: Human Language Technologies, Volume 1 (Long and Short Papers)},
  month     = jun,
  year      = {2019},
  address   = {Minneapolis, Minnesota},
  publisher = {Association for Computational Linguistics},
  url       = {https://aclanthology.org/N19-1245},
  doi       = {10.18653/v1/N19-1245},
  pages     = {2357--2367}
}

@article{bib:ArcChallenge,
  author  = {Peter Clark  and Isaac Cowhey and Oren Etzioni and Tushar Khot and
             Ashish Sabharwal and Carissa Schoenick and Oyvind Tafjord},
  title   = {Think you have Solved Question Answering? Try ARC, the AI2 Reasoning Challenge},
  journal = {arXiv:1803.05457v1},
  year    = {2018}
}
